\PassOptionsToPackage{table}{xcolor}
\documentclass[]{fairmeta}

\usepackage{amsmath}
\usepackage{amssymb}
\usepackage{amsfonts}
\usepackage{mathtools}
\usepackage{amsthm}
\usepackage{comment}
\usepackage{siunitx}
\usepackage{pifont}
\usepackage{adjustbox}
\usepackage{makecell}
\usepackage{wrapfig}
\usepackage{algorithm}
\usepackage{algorithmicx}
\usepackage{algpseudocode}
\usepackage{float}
\usepackage{listings}
\usepackage{manyfoot}
\usepackage[title]{appendix}
\usepackage{mathrsfs}
\usepackage{colortbl}
\usepackage{tabularx}
\usepackage{enumitem}
\usepackage{dblfloatfix}
\usepackage{capt-of}
\usepackage{nicefrac}
\usepackage{multicol}
\usepackage{etoolbox}
\usepackage{needspace}
\usepackage{url}
\usepackage{tikz}

\newcommand{\pfdnav}{\textsc{P4D-Nav}}
\newcommand{\evolvingworldbench}{\textsc{EvoWorld-Bench}}
\newcommand{\evolvingworldnav}{\textsc{EvolvingNav}}
\newcommand{\na}{--}
\newcommand{\paperreported}[1]{#1\textsuperscript{\scriptsize PR}}
\definecolor{tblheader}{RGB}{233,238,245}
\definecolor{tblours}{RGB}{238,240,242}
\definecolor{tblreported}{RGB}{244,246,250}
\definecolor{tblnav}{RGB}{248,242,229}
\definecolor{tblmemory}{RGB}{232,241,249}
\definecolor{tblpredict}{RGB}{242,237,248}
\newcommand{\hdr}[1]{\colorbox{tblheader}{\strut #1}}
\newcommand{\homepageicon}{%
  \tikz[baseline=-0.2em,scale=0.2]{%
    \draw[line width=0.7pt] (0,0) circle (1);
    \draw[line width=0.6pt] (0,-1) arc[start angle=-90,end angle=90,x radius=0.45,y radius=1];
    \draw[line width=0.6pt] (0,-1) arc[start angle=270,end angle=90,x radius=0.45,y radius=1];
    \draw[line width=0.6pt] (-1,0)--(1,0);
    \draw[line width=0.6pt] (-0.86,0.5)--(0.86,0.5);
    \draw[line width=0.6pt] (-0.86,-0.5)--(0.86,-0.5);
  }%
}

\newcommand{\blogicon}{%
  \tikz[baseline=-0.35em,scale=0.2]{%
    \draw[line width=0.7pt,line join=round] (-0.75,-1)--(0.75,-1)--(0.75,0.55)--(0.3,1)--(-0.75,1)--cycle;
    \draw[line width=0.6pt] (0.3,1)--(0.3,0.55)--(0.75,0.55);
    \draw[line width=0.6pt] (-0.45,0.3)--(0.4,0.3);
    \draw[line width=0.6pt] (-0.45,-0.1)--(0.4,-0.1);
    \draw[line width=0.6pt] (-0.45,-0.5)--(0.15,-0.5);
  }%
}

\newcommand{\appref}[1]{%
  \begingroup
  \crefalias{section}{appendix}%
  \crefalias{subsection}{appendix}%
  \crefname{appendix}{Appendix}{Appendices}%
  \Crefname{appendix}{Appendix}{Appendices}%
  \cref{#1}%
  \endgroup
}

\title{Beyond the Remembered World: Predictive 4D Belief for Persistent Navigation in Evolving Worlds}

\author[1,2,\ast]{Mingjian Gao}
\author[1,2,\ast]{Zhaocheng Li}
\author[1,3,\ast]{Haoyang Huang}
\author[2,\ddagger]{Wenqiao Zhang}
\author[1,4,\ddagger,\dagger]{Yingjie NIU}
\author[2]{Hao Zhou}
\author[1]{Chao Li}
\author[2]{Juncheng Li}
\author[2]{Siliang Tang}
\author[2]{Yueting Zhuang}

\affiliation[1]{Deeprobotics}
\affiliation[2]{Zhejiang University}
\affiliation[3]{University of California, San Diego}
\affiliation[4]{Chinese University of Hong Kong}

\contribution[\ast]{Equal contribution}
\contribution[\ddagger]{Corresponding authors}
\contribution[\dagger]{Project lead}
\date{\today}

\abstract{
Persistent spatial memory enables embodied agents to navigate familiar environments across repeated visits. However, targets may move while unobserved, including during navigation, making remembered locations unreliable by the time an agent arrives. Despite advances in memory retrieval and state prediction, accounting for continued hidden world evolution and revising beliefs under limited visibility remain challenging.
We study Evolving-World Navigation, where agents infer target locations from intermittent observations, predict their states at inspection time, and revise beliefs using visual evidence.
We propose \textbf{\evolvingworldnav{}}, which constructs a time-indexed belief from timestamped 3D object histories through a structured persistence--relocation model. The belief distinguishes persistence at the last observed location from relocation to alternative locations and retains probability mass outside the known candidate set.
An event-driven filter propagates the current belief as time elapses, forecasts target occupancy at candidate inspection times, and incorporates new RGB-D evidence.
Negative observations downweight location hypotheses according to calibrated, visibility-conditioned detection probabilities, while evidence tracking prevents repeated use of the same observations.
A frozen, zero-shot vision--language controller uses the updated belief to choose actions and replan.
We further introduce \textbf{\evolvingworldbench{}}, a benchmark grounded in human activity traces, comprising 54 scenes and 803,680 tasks with controlled changes before and during navigation.
In simulation and real-robot experiments, \textbf{\evolvingworldnav{}} improves navigation success and search efficiency over the evaluated baselines.
Paired experiments show the clearest gains under learnable temporal patterns, while ablations demonstrate the value of preserving uncertainty and incorporating visibility-aware evidence.

\par\smallskip\noindent

\begin{tabularx}{\linewidth}{@{}c@{\hspace{0.85em}}l@{\quad}>{\raggedright\arraybackslash}X@{}}
    \makebox[1.3em][c]{\blogicon} & \textbf{Website} & \url{https://developer.deeprobotics.cn/research/articles/evolving-nav/index.html} \\[0.45em]
    \makebox[1.3em][c]{\homepageicon} & \textbf{Homepage} & \url{https://zju4embodiedai.github.io/EvolvingNav/} \\[0.45em]
    \makebox[1.3em][c]{\raisebox{-0.2em}{\includegraphics[height=1.2em]{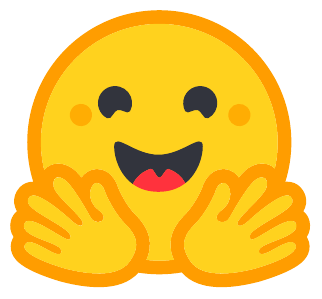}}} &
    \textbf{Demo} & \url{https://huggingface.co/spaces/ZJU4EmbodiedAI/EvolvingNav} \\[0.45em]
    \makebox[1.3em][c]{\raisebox{-0.2em}{\includegraphics[height=1.2em]{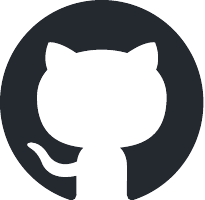}}} &
    \textbf{Code} & \url{https://github.com/ZJU4EmbodiedAI/EvolvingNav}
\end{tabularx}
}

\begin{document}
\noindent
\begin{minipage}[c][0.95cm][c]{0.62\textwidth}
    \includegraphics[height=0.82cm]{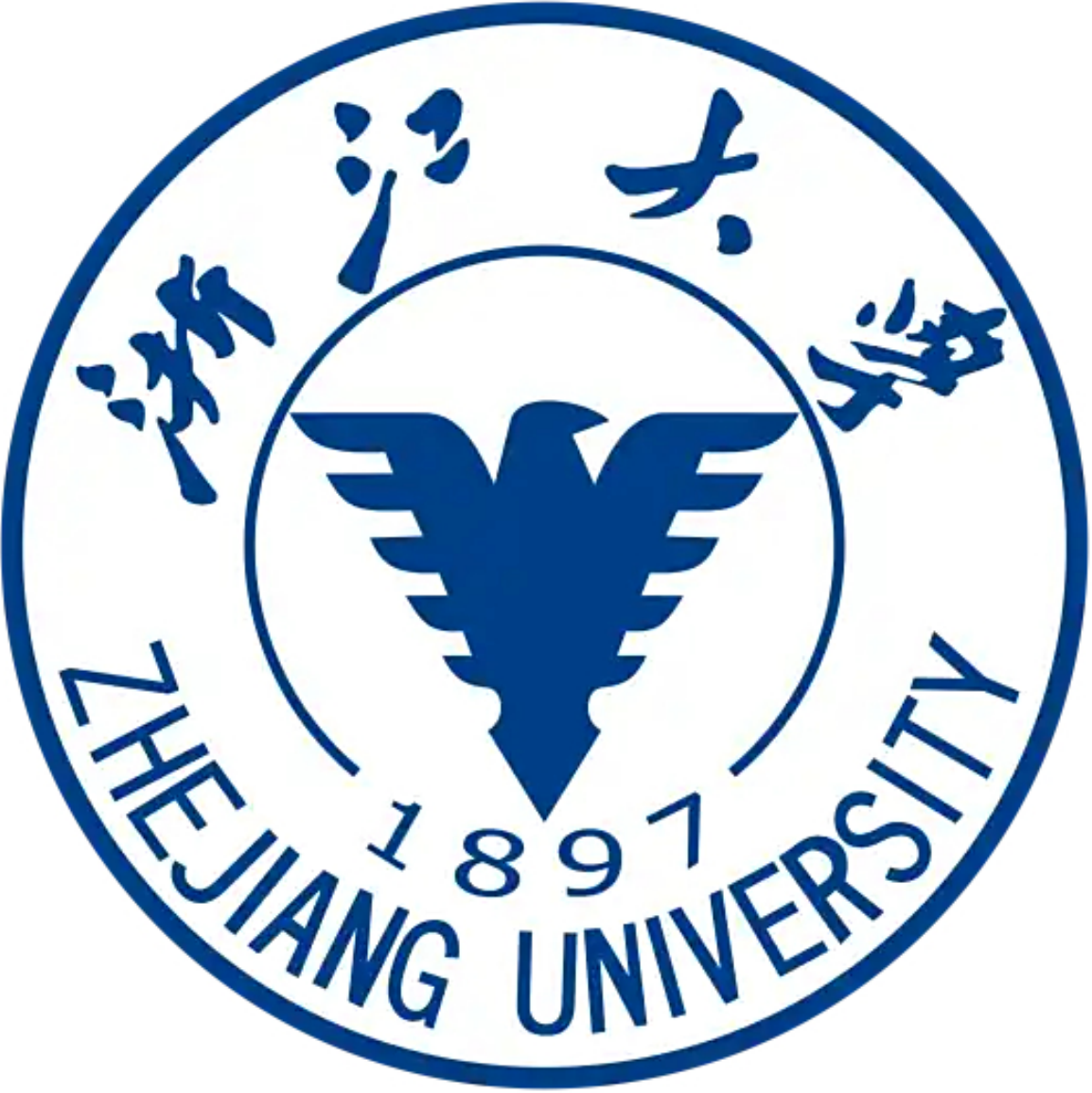}%
    \hspace{0.28cm}%
    \includegraphics[width=3.05cm]{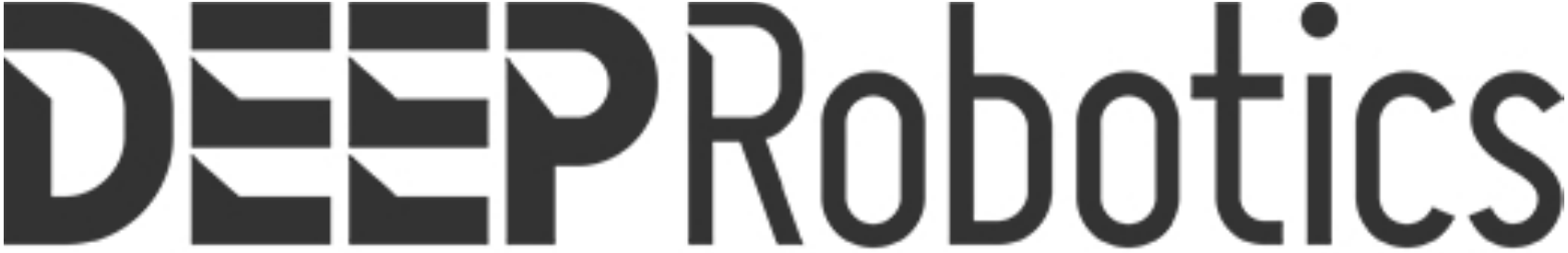}%
\end{minipage}%
\begin{minipage}[c][0.95cm][c]{0.38\textwidth}
    \raggedleft
    \includegraphics[height=0.82cm]{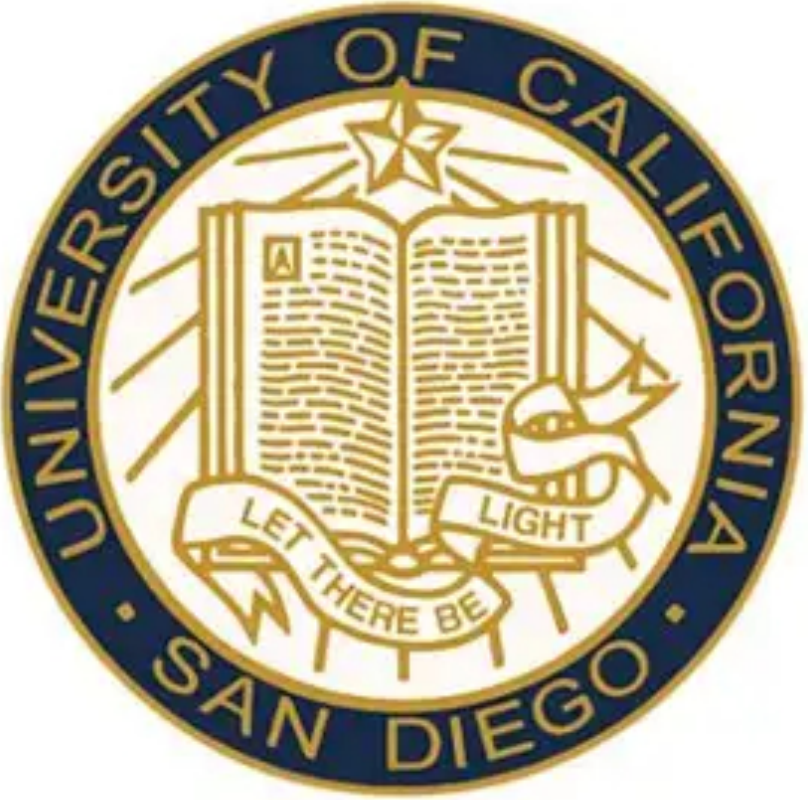}%
    \hspace{0.25cm}%
    \includegraphics[height=0.82cm]{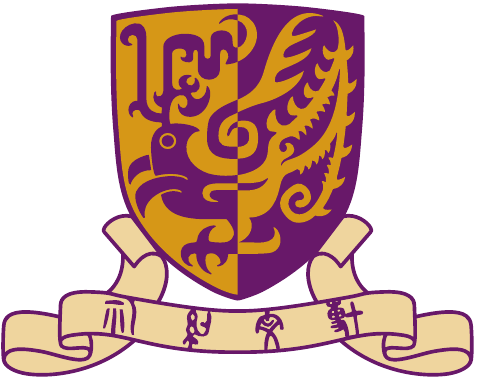}%
\end{minipage}\par
\vspace{-0.08cm}
\noindent\rule{\textwidth}{0.4pt}\par
\vspace{0.08cm}
\maketitle
\raggedbottom

\begin{figure}[t]
    \centering
    \includegraphics[width=\textwidth]{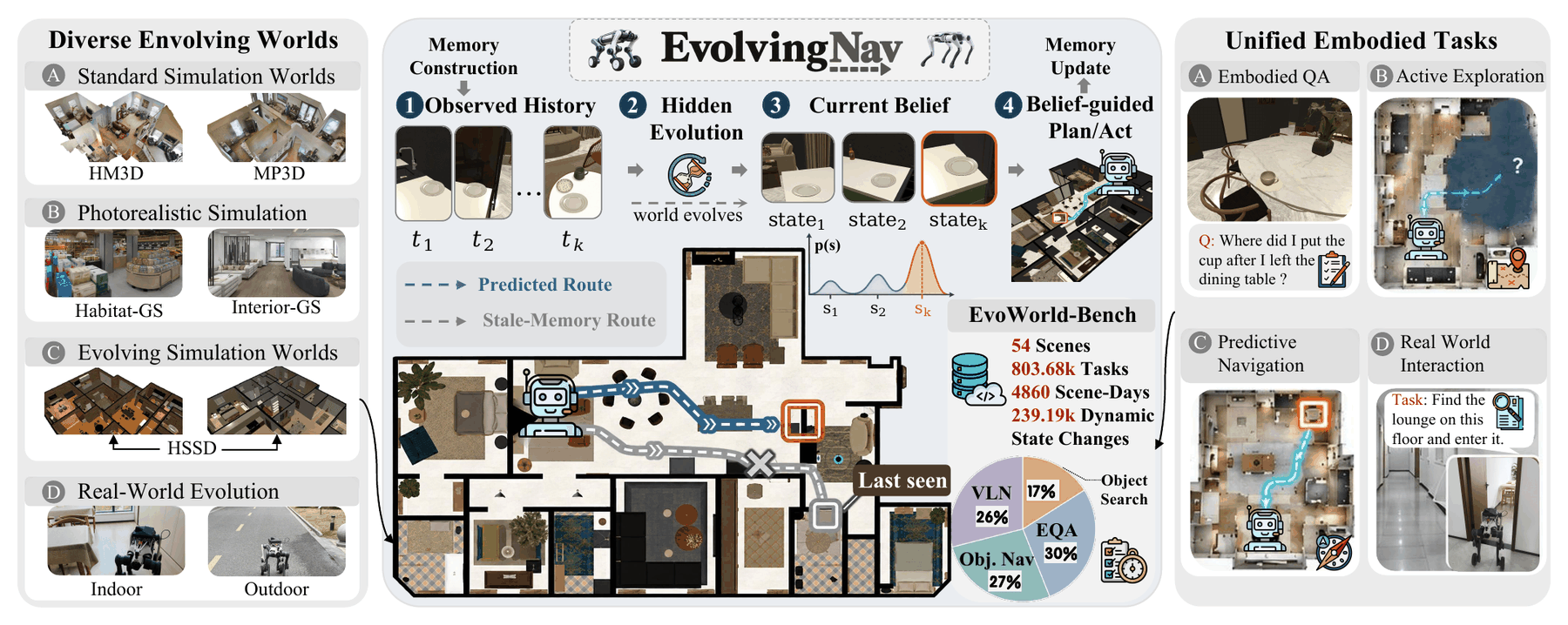}
    \caption{Overview of \evolvingworldnav{} for predictive embodied reasoning and action in evolving worlds.}
    \label{fig:evolvingworldnav_overview}
\end{figure}

\section{Introduction}
\label{sec:introduction}
Long-lived embodied agents navigate familiar environments across repeated tasks. 
Spatial maps and object memories preserve observations from earlier visits \citep{gu2024conceptgraphs,liu2024dynamem}. 
However, those observations may not describe the world when the agent acts. 
Human activity changes task-relevant states outside observation: a medication box observed on a bedside table may return to a cabinet before the next request. 
Further changes can occur while the robot travels, invalidating a prediction made at the start of navigation. 
The agent must therefore plan from observations acquired minutes, hours, or days earlier and estimate what will be true when it inspects a target.

Under partial observability, historical observations are evidence about past states, while the current state remains latent and may evolve. 
An accurate map can therefore lead an agent to an outdated site, while searching from scratch discards useful history. 
A time-indexed predictive belief should represent uncertainty over unobserved relocations and support revision during interaction. 
Candidate destinations have different travel times, so their occupancy probabilities must be evaluated at their respective arrival times. 
A failed inspection should weaken a location hypothesis only when the site was adequately visible. 
If changes continue during execution, even an inspected site may later become plausible again.

Existing work has studied both temporal prediction and navigation with moving objects. 
PredictiveGraphs predicts future object--receptacle states, ranks candidate destinations, and updates its estimates during navigation \citep{saavedraruiz2026predictivegraphs}. 
Transit-Aware Planning considers portable targets that move while an agent travels \citep{dorbala2026portable}. 
We focus on combining irregular observation histories, candidate-specific arrival times, and visibility-conditioned evidence within one navigation loop. The agent must also retain probability mass outside its known candidate set.


We formalize this problem as \textbf{Evolving-World Navigation}. 
Given a target, a timestamped observation history, candidate locations, and an action budget, the agent must locate and visually verify the target without access to hidden transitions or current ground truth. 
It receives new observations only along its executed path. 
Portable-object search provides a concrete instance because people can move objects outside the robot's view. 
We consider two episode types. 
In fixed-target episodes, a target may move before the query but remains fixed during the search. 
In continuing-evolution episodes, it may move while the agent navigates.


We introduce \textbf{\evolvingworldnav{}} (\cref{fig:evolvingworldnav_overview}), a navigation agent built around a predictive 4D belief. 
Its persistent memory records entity identity, 3D spatial context, observation times, and evidence provenance. Using irregularly sampled histories, \pfdnav{} predicts whether the target is at its last observed site, at another known site, or somewhere outside the known candidate set. 
It assigns probability to all three possibilities, keeping alternatives available as the robot gathers new evidence. 
A frozen, zero-shot vision--language controller uses the belief to call memory, prediction, inspection, exploration, and navigation tools.

The agent closes the loop with an event-driven predict--observe--replan filter. 
Before choosing a destination, it predicts target occupancy at each candidate's estimated arrival time and weighs that prediction against expected new coverage and travel cost. 
It then moves in short segments and updates the belief using RGB-D observations gathered along the way and at inspection sites. 
A clear view of an empty site can redirect its route; an occluded view leaves the corresponding hypothesis largely intact. 
Time-valid evidence rounds prevent correlated frames from being counted repeatedly. 
As time passes, the model can also restore probability to a site inspected earlier.


We introduce \textbf{\evolvingworldbench{}} to evaluate these decisions under changing conditions. 
Grounded in human trajectories, it contains 54 evolving scenes and 803,680 task instances spanning state prediction, object navigation, vision-and-language navigation, and embodied question answering. 
The benchmark measures how predictions affect first inspection, recovery, path efficiency, and online replanning. Paired static, routine, and random worlds keep scenes and queries fixed while varying temporal structure. 
These comparisons test when learnable patterns help an agent predict and search.

Across the navigation suite, \textbf{\evolvingworldnav{}} improves initial inspection and eventual search (\cref{tab:crossbench}). 
When targets can move during execution, it recovers more reliably and avoids unnecessary travel (\cref{tab:n4-online}). 
Paired experiments show the clearest benefit when changes follow learnable routines. 
Ablations examine the roles of alternative hypotheses and visibility-aware updates. 
Held-out scenes, cross-benchmark tests, different frozen VLMs, and physical-robot trials assess performance across the evaluated settings.

Our contributions are threefold:
\begin{itemize}[leftmargin=*,itemsep=0.6mm,topsep=1mm]
    \item \textbf{Problem setting:} We study navigation with hidden changes before a query and, in continuing-evolution episodes, during execution. 
    The agent must reason about arrival times and visually verify the target.
    \item \textbf{Belief-driven agent:} We introduce \textbf{\evolvingworldnav{}}, combining a time-indexed belief over known and unknown locations with an event-driven predict--observe--replan filter.
    \item \textbf{Benchmark and evaluation:} We introduce \textbf{\evolvingworldbench{}} and use paired temporal controls, simulation, and physical-robot trials to evaluate prediction, evidence-based recovery, and navigation efficiency.
\end{itemize}

\section{Related Work}
\label{sec:related_work}

\paragraph{Navigation, memory, and changing worlds.}
Embodied navigation combines perception, spatial reasoning, and action;
open-vocabulary maps and scene graphs ground language in spatial representations
\citep{anderson2018evaluation,savva2019habitat,chaplot2020objectnav,
xin2026agentvln,huang2023vlmaps,gu2024conceptgraphs,werby2024hovsg,
Li_2026_CVPR,huang2026msgnav}. Foundation-model agents combine maps, tools, and
navigation skills
\citep{yokoyama2024vlfm,long2024instructnav,ziliotto2025tango,
li2026agenticnav,zheng2021cos,ginting2024seek,zhang2025apexnav}. Lifelong
benchmarks evaluate navigation and memory across tasks
\citep{khanna2024goatbench,yadav2025findingdory}, while embodied QA benchmarks
assess question answering from episodic observations
\citep{majumdar2024openeqa}. Dynamic maps and 4D memories support updates and
spatiotemporal retrieval
\citep{sohn2025r4,gorlo2026describe,bescos2018dynaslam,
narayana2020lifelong,liu2024dynamem}. EmbodiedSkills verifies skill
execution in a closed-loop VLA agent, while VisualThink-VLA routes compact
visual evidence for robot control
\citep{wang2026embodiedskills,gao2026visualthinkvla}.
Complementary multimodal work studies interleaved visual--language
instructions (Cheetah), adaptive visual and language experts (HyperLLaVA),
instruction curation (Align$^2$LLaVA), and instruction-driven instance
segmentation (InstructSAM)
\citep{li2024cheetah,zhang2024hyperllava,huang2024align2llava,
yuan2026instructsam}.

\paragraph{Predictive memory and human routines.}
Predictive navigation models infer object locations from partial histories
using discrete states or continuous spatial representations
\citep{kurenkov2023dynamic,saavedraruiz2026predictivegraphs,
argenziano2026flowmaps}. PredictiveGraphs combines future-state
prediction, Top-$k$ navigation, online observations, and agent tools; our
focus is an arrival-time filtering loop for continued hidden
evolution, with time-valid evidence and candidate reopening. HOMER/HOMER+,
STREAK, and personalized navigation model household routines or context drift
\citep{patel2023homer,patel2023routine,bartoli2025streak,zhang2026ucon}.
HD-EPIC and ParaHome provide human--object interaction traces
\citep{perrett2025hdepic,kim2025parahome}. \evolvingworldbench{} uses human traces and
separately tracked simulated evidence to constrain executable histories; paired
worlds test temporal regularity beyond location frequency.

\section{Predictive 4D Belief Navigation}
\label{sec:method}

\begin{figure}[t]
    \centering
    \includegraphics[width=\linewidth]{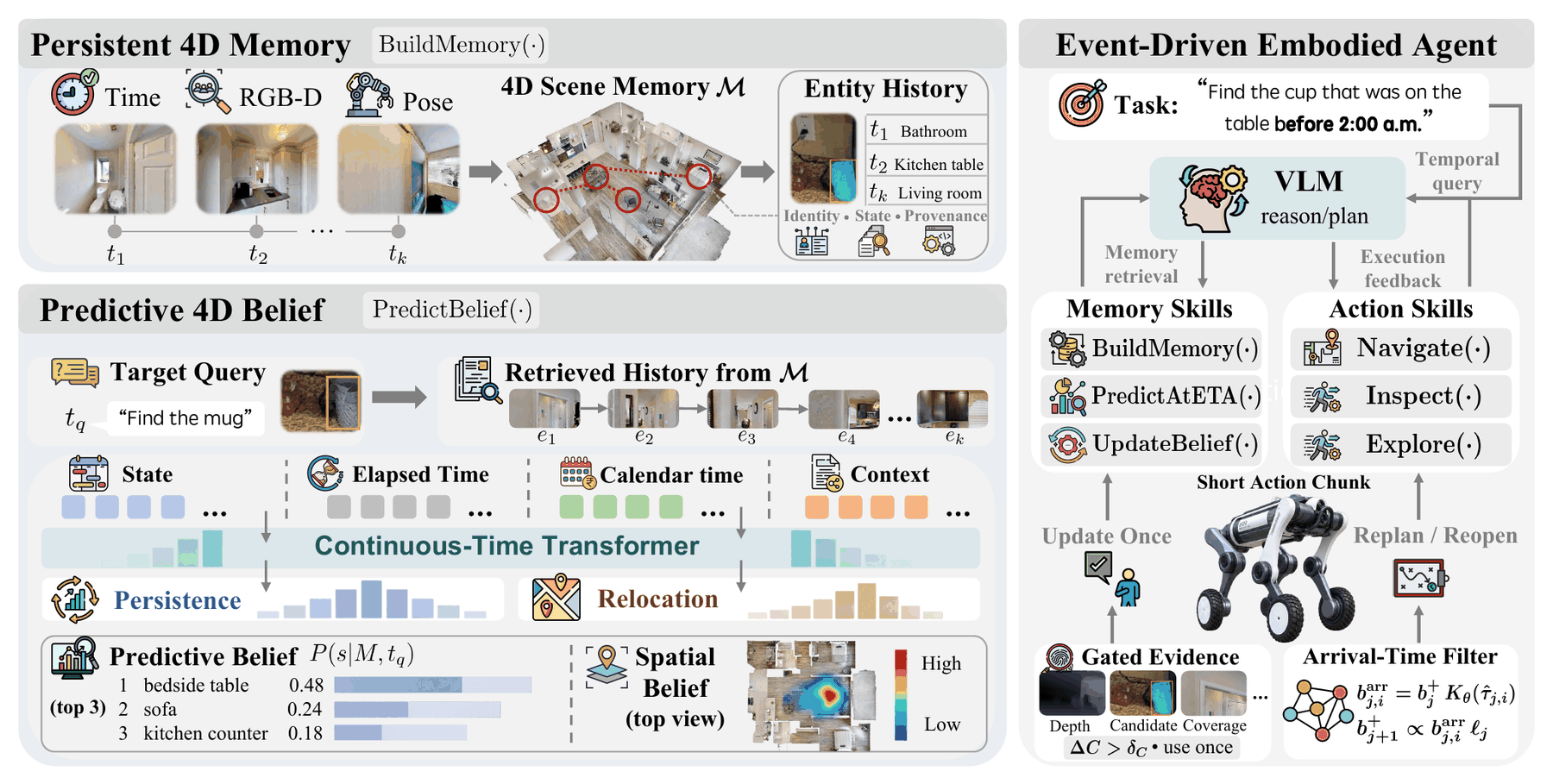}
    \caption{\evolvingworldnav{} couples persistent 4D memory, predictive belief,
    and an embodied agent that updates and replans from stepwise evidence.}
    \label{fig:p4dnav_method}
\end{figure}

\evolvingworldnav{} maintains a current-state belief from timestamped 3D histories
and RGB-D evidence; \pfdnav{} separates persistence from relocation
(\cref{fig:p4dnav_method}).

\subsection{Problem Formulation}
\label{sec:problem}

An environment contains a navigable space $\mathcal X$, entities $\mathcal O$,
and candidate states $\mathcal C_o$ with inspection viewpoints. Before query
time $t_q$, the agent has only the causal history
\begin{equation}
\mathcal H_{\leq t_q}=\{(I_k,D_k,\xi_k,t_k)\mid t_k\leq t_q\},
\end{equation}
where $I_k,D_k,\xi_k$ denote RGB, depth, and camera pose. The hidden state lies in
$\widetilde{\mathcal C}_o=\mathcal C_o\cup\{\mathrm{unknown}\}$; later
out-of-view transitions are also hidden.

Given a target, initial pose $x_0$, memory, and action budget, the agent observes
$z_j$ after acting and must stop at a valid target viewpoint. It optimizes
\begin{equation}
\pi^*=\arg\max_{\pi}\;\mathbb E_{\pi}\!\left[
\mathbb I(\mathrm{success})-\lambda_d L-\lambda_n N_{\mathrm{inspect}}
\right],
\label{eq:objective}
\end{equation}
where $L$ is the travel distance and $N_{\mathrm{inspect}}$ is the inspection count.

\subsection{Causal 4D Memory and History Encoding}
\label{sec:memory_encoding}
The 4D memory registers RGB-D observations in a shared frame, associates entities using
semantics, geometry, and time, and materializes versioned deltas causally:
\begin{equation}
\mathcal M_{\leq t}=\operatorname{Materialize}(\Delta\mathcal M_0,\ldots,\Delta\mathcal M_t)
=(\mathcal G_t,\mathcal V_t,\mathcal B_t),
\label{eq:memory}
\end{equation}
where $\mathcal G_t,\mathcal V_t,\mathcal B_t$ store relations, time-valid
versions, and provenance. Changes append versions without erasing history;
perception and versioning details are in \appref{app:memory-details}
\citep{liu2024grounding,ravi2025sam,radford2021learning}.

\paragraph{Target-conditioned history and continuous time.}
For target $o$, a typed query returns the most recent $K$ causal events
\begin{equation}
H_o=\{e_k=(o_k,s_k,t_k,z_k,c_k,v_k)\}_{k=1}^{K},
\label{eq:history}
\end{equation}
where $o_k,s_k,t_k,z_k$ encode entity, candidate state, time, and
$\{\mathrm{seen},\mathrm{notseen}\}$ evidence; $c_k,v_k$ encode confidence and
visibility. Related context events are causal, candidate encodings capture
function and geometry, and \texttt{unknown} absorbs unmapped states.

Each event token combines semantic, evidence, and temporal features:
\begin{align}
\mathbf x_k={}&E_o(o_k)+E_s(s_k)+E_z(z_k)+E_c(c_k)+E_v(v_k)\nonumber\\
&+\phi_{\Delta}(t_q-t_k)+\phi_{\mathrm{cal}}(t_k),
\label{eq:eventtoken}
\end{align}
where $\phi_{\Delta}$ encodes log elapsed time and $\phi_{\mathrm{cal}}$ encodes
hour and weekday. A query token specifies the target, time, and last observed state; a compact
Transformer yields
\begin{equation}
(\mathbf h_1,\ldots,\mathbf h_K,\mathbf h_q)
=\operatorname{CTTransformer}(\mathbf x_1,\ldots,\mathbf x_K,\mathbf x_q).
\label{eq:history-encoder}
\end{equation}
The encoding captures irregular intervals, evidence, and event order without hidden labels.

\subsection{Persistence--Relocation Belief}
\label{sec:belief_model}
We predict persistence at the last observed state or relocation elsewhere.
Episode validity guarantees a last positively observed state $s_{\mathrm{last}}$
(\appref{app:audit}). The persistence branch predicts
\begin{equation}
\rho_q=P(s_{t_q}^o=s_{\mathrm{last}}\mid H_o,t_q)
=\sigma(\mathbf w_{\rho}^{\top}\mathbf h_q).
\label{eq:persistence}
\end{equation}
For each alternative $i\in\widetilde{\mathcal C}_o\setminus
\{s_{\mathrm{last}}\}$, a shared pointer head computes
\begin{equation}
a_i=\frac{(W_q\mathbf h_q)^\top(W_c\mathbf g_i)}{\sqrt d}
+\psi(\mathbf h_q,\mathbf g_i),\qquad
\pi_q(i)=\frac{\exp a_i}{\sum_{u\in\widetilde{\mathcal C}_o\setminus\{s_{\mathrm{last}}\}}\exp a_u},
\label{eq:pointer}
\end{equation}
where $\mathbf g_i$ represents candidate $i$, $d$ is the projection dimension,
and $\psi$ measures learned compatibility. The prior is
\begin{equation}
b_q^-(i)=
\begin{cases}
\rho_q, & i=s_{\mathrm{last}},\\
(1-\rho_q)\pi_q(i), & i\neq s_{\mathrm{last}}.
\end{cases}
\label{eq:prior}
\end{equation}
We train with $\mathcal L_{\mathrm{state}}=-\log b_q^-(s_{t_q}^o)$ without
change labels; the normalized belief represents returns and alternatives.

\subsection{Predictive Embodied Agent}
\label{sec:agent}

A frozen VLM invokes memory, prediction, navigation, inspection, and exploration
tools. Action chunks, new coverage or evidence, ETA changes, and arrivals trigger
decision epochs. The encoder produces $\mathbf h_j$ for the row-normalized operator
\begin{equation}
[K_\theta^{(j)}(\Delta t)]_{r,u}=\operatorname{softmax}_{u}
f_\theta(\mathbf h_j,\mathbf g_r,\mathbf g_u,\phi_\Delta(\Delta t)).
\label{eq:transition-kernel}
\end{equation}
The operator shares the encoders in \cref{eq:history-encoder,eq:pointer}; a separate head
learns chronological state transitions through row-wise cross-entropy
(\appref{app:belief-control}). Initialized with $b_0^-=b_q^-$, the filter propagates after each chunk:
\begin{equation}
b_{j+1}^-=b_j^+K_\theta^{(j)}(t_{j+1}-t_j).
\label{eq:filter-predict}
\end{equation}
Thus $b_j^\pm$ denotes the current-time belief, separate from arrival forecasts.

\paragraph{Arrival-aware action selection.}
For candidate viewpoints $\mathcal W_i$, let
$d_{j,i}=d_{\mathrm{geo}}(x_j,\mathcal W_i)$ and $\widehat\tau_{j,i}$ be
the geodesic distance and estimated travel-plus-inspection time. The agent selects
\begin{equation}
i_j^*=\arg\max_{i\in\mathcal{C}_o}
\frac{
    p_{j,i}^{\mathrm{arr}}\,
    \widehat{q}_{j,i}^{\,\mathrm{new}}
}{
    d_{j,i}
    + \lambda_t \widehat{\tau}_{j,i}
    + \lambda_{\mathrm{inspect}}
},
\label{eq:planner}
\end{equation}
where $p_{j,i}^{\mathrm{arr}}$ is defined below and
$\widehat q_{j,i}^{\mathrm{new}}$ is the detection probability from newly covered
surfaces. Exploration expands the candidate set when unknown-state utility is highest.

\paragraph{Visibility-qualified measurement update.}
A posed RGB-D view can update every covered candidate. A calibrated detector estimates
$\widehat r_{j,i}=P(\mathrm{detect}\mid s_{t_j}=i,\mathbf f_{j,i})$ from
projected candidate geometry, online depth, and view features, without access to
ground-truth target masks or poses or simulator visibility flags. Given no detection,
\begin{equation}
\ell_j(i)=P(y_j=\varnothing\mid s_{t_j}=i)=
\begin{cases}
1-\widehat r_{j,i}, & i\in\mathcal C_o,\\
1, & i=\mathrm{unknown},
\end{cases}
\label{eq:negative}
\end{equation}
and the measurement posterior is
\begin{equation}
b_j^+(i)=\frac{b_j^-(i)\ell_j(i)}
{\sum_{u\in\widetilde{\mathcal C}_o}b_j^-(u)\ell_j(u)}.
\label{eq:bayes}
\end{equation}
An update uses only new surface coverage exceeding $\delta_C=0.05$, preventing
duplicate evidence from overlapping views. Dynamic reopening starts a new
time-valid round, allowing the same viewpoint to test a changed state. A verified match ends the search.

\paragraph{Posterior propagation and replanning.}
The current posterior is forecast to each candidate ETA for action selection,
without replacing the current-time filter:
\begin{equation}
b_{j,i}^{\mathrm{arr}}=b_j^+K_\theta^{(j)}(\widehat\tau_{j,i}),\qquad
p_{j,i}^{\mathrm{arr}}=b_{j,i}^{\mathrm{arr}}(i).
\label{eq:arrival-belief}
\end{equation}
Fixed-target episodes use identity dynamics; continuing evolution follows
\cref{eq:filter-predict,eq:arrival-belief}. At each epoch, the agent propagates
belief, incorporates new evidence once, forecasts arrival-time states, and replans
(\appref{app:belief-control}).

\section{EvoWorld-Bench: Evaluating Navigation in Evolving Worlds}
\label{sec:benchmark}

\evolvingworldbench{} evaluates decisions based on time-ordered histories while the
world changes outside observation. Its 54 scenes and 803.68k task instances
combine persistent histories, executable placements, and controlled dynamics.

\subsection{Tasks and Controlled Dynamics}

At the task-modality level in \cref{fig:evolvingworldnav_overview}, the pool
comprises object search (17\%), object navigation (27\%), vision-and-language
navigation (VLN; 26\%), and embodied question answering (EQA; 30\%)
\citep{majumdar2024openeqa,sakamoto2024map}. These labels describe task goals;
the protocols below distinguish prediction, search, evidence, and online dynamics.

\Needspace{0.20\textheight}
\begin{wrapfigure}{r}{0.58\textwidth}
    \centering
    \vspace{-8pt}
    \includegraphics[width=\linewidth]{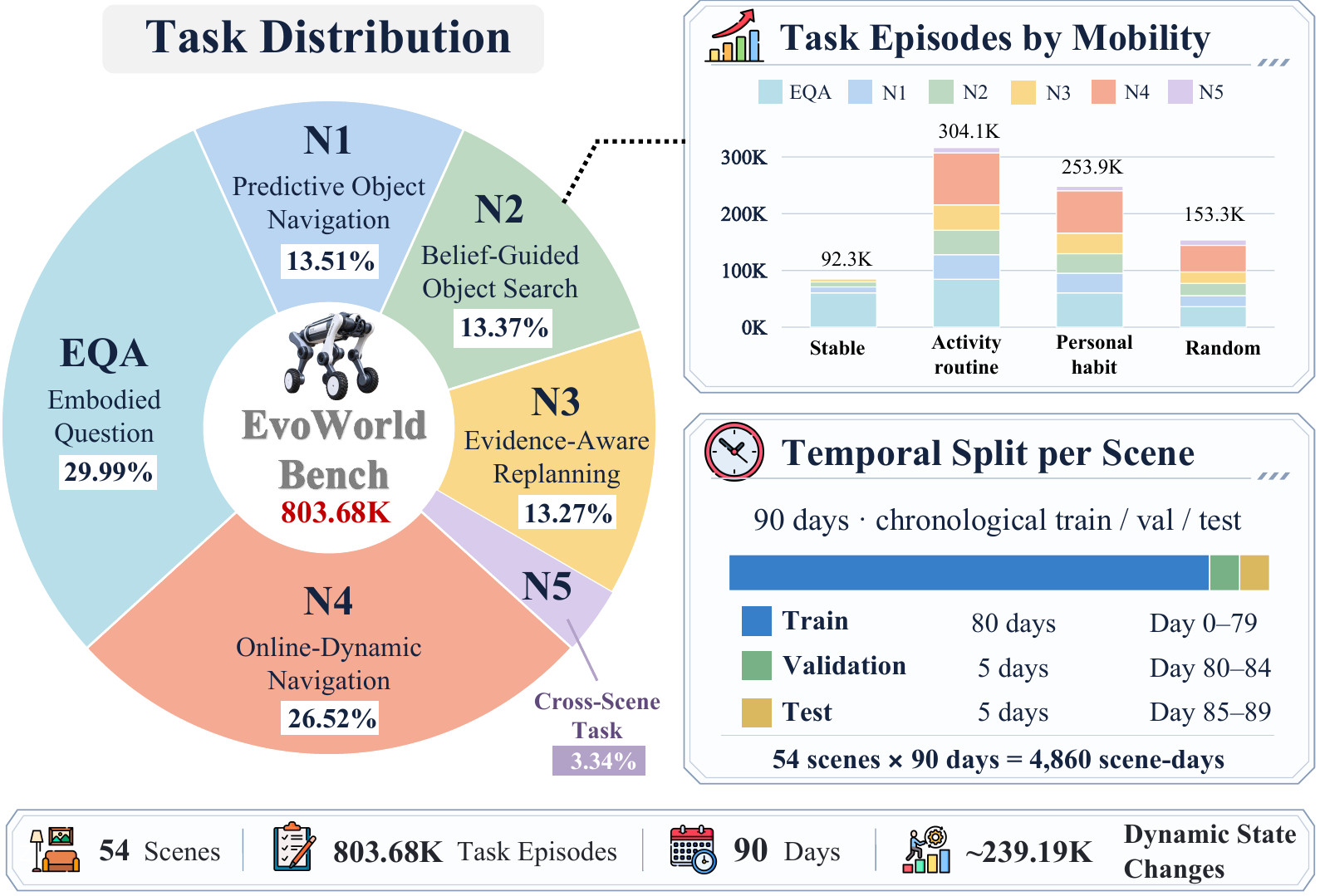}
    \caption{Overview of \evolvingworldbench{} evaluation protocols.}
    \label{fig:p4dbench}
\end{wrapfigure}

The complementary breakdown by evaluation protocol in \cref{fig:p4dbench}
comprises N1 (13.51\%), N2
(13.37\%), N3 (13.27\%), N4 (26.52\%), N5 (3.34\%), and EQA (29.99\%). In this
second view, VLN-conditioned episodes are assigned to their corresponding
navigation protocol rather than a separate category. Prediction evaluates the
region/receptacle belief independently of control; navigation uses the same
target--time queries (\cref{fig:p4dbench,tab:tasks}). N1 inspects the Top-1
destination; N2 uses the full belief for cost-aware search; N3 adds
visibility-qualified updates and recovery. N1--N3 fix the target after the
query. N4 advances the world clock during execution, permitting further hidden
transitions. N5 applies these protocols to held-out scenes and households.

VLN grounds language goals in the same histories; EQA probes current and
historical states without future evidence. Our experiments emphasize prediction
and navigation, with EQA as a memory diagnostic. Controlled comparisons match public
observations, candidates, geometry, starts, controllers, and budgets.

Paired \emph{static}, \emph{routine}, and \emph{random} worlds share public setup
variables. Static preserves the last supported state; routine follows household
dynamics; random matches legal candidates and marginal movement statistics while
removing temporal dependencies. Improvements specific to routine worlds thus
support the use of historical regularity over category frequency or spatial
convenience.

\subsection{World Construction and Past-Only Evaluation}
Human traces constrain generation; source trajectories are not copied
(\cref{fig:p4dbench-generation}). CASAS supplies longitudinal occupancy and
timing statistics; ARAS and OPPORTUNITY add activity-context coverage; and
HD-EPIC and ParaHome provide object--action and motion evidence
\citep{cook2013casas,perrett2025hdepic,kim2025parahome}. HOMER+ is tracked
separately as simulated long-horizon routine evidence \citep{patel2023routine}.
Event records preserve source identifiers and distinguish measured,
environment-derived, simulated, and benchmark-authored quantities
(\appref{app:benchmark-construction}).

Across 54 HSSD scenes \citep{khanna2024hssd}, household habits and stochastic
exceptions generate chronological histories. Stable, routine,
personal, and irregular regimes test persistence, activity regularity,
owner-specific habits, and uncertainty (\cref{tab:mobility-regimes}); regime labels
remain hidden from the agent. Habitat replay ensures semantically valid, collision-free placements
and reachable inspection viewpoints.

\noindent\begin{minipage}[t]{0.47\linewidth}
\vspace{0pt}
Patrols collect visibility-qualified RGB-D observations using only maps, time, travel cost,
and observed history. Queries expose only observations available by the query
time, the last supported
state, elapsed/calendar time, candidates, and spatial context; current states,
transitions, activities, mobility labels, and oracle viewpoints remain private.
Splits group complete temporal sequences, paired worlds, and near-duplicate
query/transition groups. Audits cover temporal and metadata leakage, split overlap,
paired-world consistency, physical validity, and patrol access to hidden state.
Stratification covers mobility, transition, staleness, spatial difficulty, and
generalization.
\end{minipage}\hfill
\begin{minipage}[t]{0.50\linewidth}
\vspace{0pt}
\centering
\includegraphics[width=\linewidth]{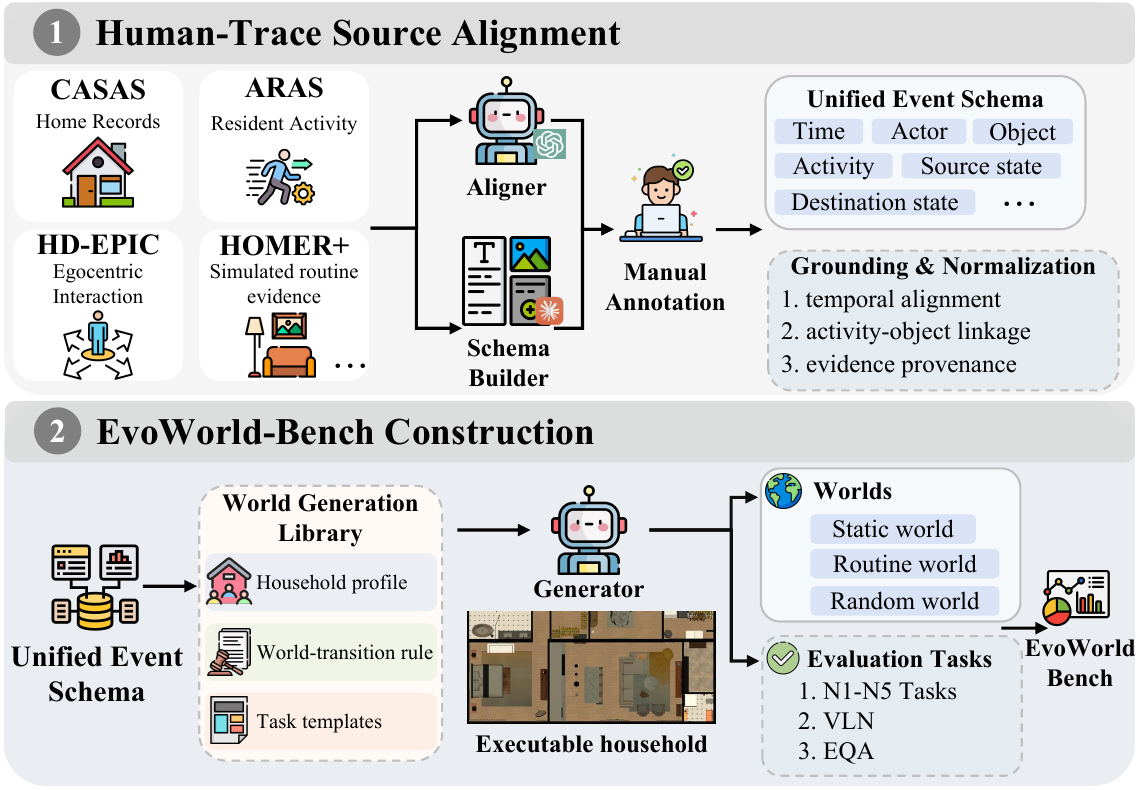}
\captionof{figure}{Construction of \evolvingworldbench{}.}
\label{fig:p4dbench-generation}
\end{minipage}\par\medskip

\section{Experiments}
\label{sec:experiments}

We test whether predictive 4D memory improves current-state inference,
embodied search, and recovery when observations contradict prior beliefs. Controlled
\evolvingworldbench{} and robot comparisons use the same public histories, legal candidates,
inspection viewpoints, low-level control, and action budgets. Cross-environment
evaluations retain the native policies of released agents and provide
system-level generalization evidence. Component ablations support causal
comparisons. Further protocol, implementation, and analysis details are
summarized in Appendices~A--D.

\subsection{Setup}
\label{sec:setup}

\paragraph{Datasets.}
Our primary evaluation uses the complete \evolvingworldbench{} N1--N5 suite for
First-Inspection navigation, belief-guided search, evidence-aware replanning,
online dynamics, and cross-scene transfer. FindingDory
\citep{yadav2025findingdory} and GOAT-Bench \citep{khanna2024goatbench} test
long-horizon memory and repeated-goal navigation. We additionally evaluate one
native task in each of MP3D, HM3D, Habitat-GS, and InteriorGS, and run 64
matched LYNX M20 trials per method.

\paragraph{Real-world setup.}
We use the DEEP Robotics LYNX M20 for the primary quantitative
evaluation over 64 matched trials, with X30 and Lite3 evaluated on matched
32-block transfer subsets. Platform-specific perception and low-level
navigation use a common interface, and all methods receive
identical pre-query histories, candidate locations, start conditions, and
robot-specific control stacks. Each 360-second episode requires online
target confirmation within three inspections.

\paragraph{Controllers and baselines.}
Our main method uses a frozen GPT-5.6-Luna controller in a zero-shot setting:
the VLM receives no navigation-task fine-tuning and invokes the fixed memory,
prediction, inspection, exploration, and navigation tools. The prompt, tool
interface, and inference budget remain fixed within each comparison. Baselines
span navigation, structured memory, and predictive world models; paired runs
with GPT-4o, GPT-5.5, GPT-5.6-Luna, and Qwen2.5-VL-3B/32B test controller
dependence.

\paragraph{Metrics.}
We evaluate prediction using Top-1, MRR, NLL, ECE, and true-state rank, and navigation using SR,
SPL, First-Inspection SR, Recovery SR, inspection count, distance, and time. Main
comparisons report five-seed variation; paired static, routine, and random
worlds isolate learnable temporal structure. 

\subsection{Main Results}
\label{sec:results}

\paragraph{Comparison across embodied benchmarks.}

\begin{table*}[!htbp]
\caption{Cross-benchmark comparison (\%). 
$\pm$ denotes standard deviation over five seeds.}
\label{tab:crossbench}
\centering
\scriptsize
\setlength{\tabcolsep}{4.0pt}
\resizebox{\textwidth}{!}{%
\begin{tabular}{lcccccccc}
\toprule
& \multicolumn{2}{c}{\textbf{FindingDory}} &
\multicolumn{3}{c}{\textbf{GOAT-Bench}} &
\multicolumn{3}{c}{\textbf{\evolvingworldbench{}}}\\
\cmidrule(lr){2-3}\cmidrule(lr){4-6}\cmidrule(lr){7-9}
\textbf{Method} & \textbf{HL-SR $\uparrow$} & \textbf{HL-SPL $\uparrow$} &
\textbf{SR $\uparrow$} & \textbf{SPL $\uparrow$} &
\textbf{Repeat-SR $\uparrow$} & \textbf{First-Inspection SR $\uparrow$} &
\textbf{Search SR $\uparrow$} & \textbf{SPL $\uparrow$}\\
\midrule
\rowcolor{tblnav}\multicolumn{9}{c}{\emph{Open-vocabulary and lifelong navigation}}\\
VLFM \citep{yokoyama2024vlfm} & 13.42 & 10.86 & 15.79 & 5.45 & 11.82 & 20.65 & 28.36 & 24.67\\
ZSON \citep{majumdar2022zson} & 27.89 & 21.81 & 29.64 & 2.45 & 16.37 & 26.84 & 32.13 & 28.55\\
FindingDory Agent \citep{yadav2025findingdory} & \textbf{\paperreported{52.44}} & \textbf{40.92} & \na & \na & \na & \na & \na & \na\\
LagMemo GLUE \citep{zhou2025lagmemo} & 32.79 & 23.25 & 20.61 & 1.74 & 11.27 & 30.18 & 42.27 & 33.82\\
\midrule
\rowcolor{tblmemory}\multicolumn{9}{c}{\emph{Structured spatio-temporal memory}}\\
HOV-SG \citep{werby2024hovsg} & 21.47 & 15.20 & 28.37 & 2.23 & 12.38 & 36.46 & 52.32 & 40.61\\
DynaMem \citep{liu2024dynamem} & 30.30 & 22.78 & 14.54 & 1.71 & 15.79 & 45.33 & 71.94 & 56.83\\
\midrule
\rowcolor{tblpredict}\multicolumn{9}{c}{\emph{Predictive world and state modeling}}\\
SLaTe-PRO \citep{patel2023routine} & 18.18 & 14.53 & 9.41 & 0.99 & 6.38 & 13.00 & 36.33 & 26.37\\
SGM+NEP \citep{kurenkov2023dynamic} & 34.20 & 19.48 & 10.67 & 1.64 & 2.56 & 27.26 & 41.82 & 37.00\\
FlowMaps \citep{argenziano2026flowmaps} & 13.03 & 10.24 & 25.07 & 1.32 & 7.25 & 14.33 & 32.67 & 17.35\\
PredictiveGraphs \citep{saavedraruiz2026predictivegraphs} & 24.24 & 13.50 & 10.27 & 1.80 & 7.48 & 36.18 & 43.67 & 40.36\\
\midrule
\rowcolor{tblours} & \textbf{53.22} & \textbf{38.83} &
\textbf{35.43} & \textbf{12.98} & \textbf{19.31} & \textbf{61.32} &
\textbf{86.18} & \textbf{70.15}\\
\rowcolor{tblours}\multirow{-2}{*}{\textbf{\evolvingworldnav{} (ours)}} & $\pm 3.87$ & $\pm 2.43$ & $\pm 2.16$ &
$\pm 0.92$ & $\pm 0.68$ & $\pm 3.235$ & $\pm 2.07$ & $\pm 1.74$\\
\bottomrule
\end{tabular}
}
\vspace{1pt}

\parbox{\textwidth}{\scriptsize
\textsuperscript{PR} Paper-reported under the cited native protocol; unmarked
baseline scores are our reproductions under the method-preserving protocol in
\appref{app:experimental-setup}.}
\end{table*}

\evolvingworldnav{} performs strongly across FindingDory, GOAT-Bench, and
\evolvingworldbench{} (\cref{tab:crossbench}). On \evolvingworldbench{} N1--N5,
First-Inspection SR increases from 45.33\% (DynaMem) to 61.32\%, and Search SR
from 71.94\% to 86.18\%. On FindingDory, it achieves the highest HL-SR
(53.22\%) among the listed methods and the second-highest HL-SPL (38.83\%),
behind FindingDory Agent (40.92\%). The gains on \evolvingworldbench{} show the benefit of predictive belief under
hidden world evolution; results on FindingDory and GOAT-Bench indicate broader
applicability to embodied navigation.

\begin{figure}[!t]
    \centering
    \includegraphics[width=\linewidth]{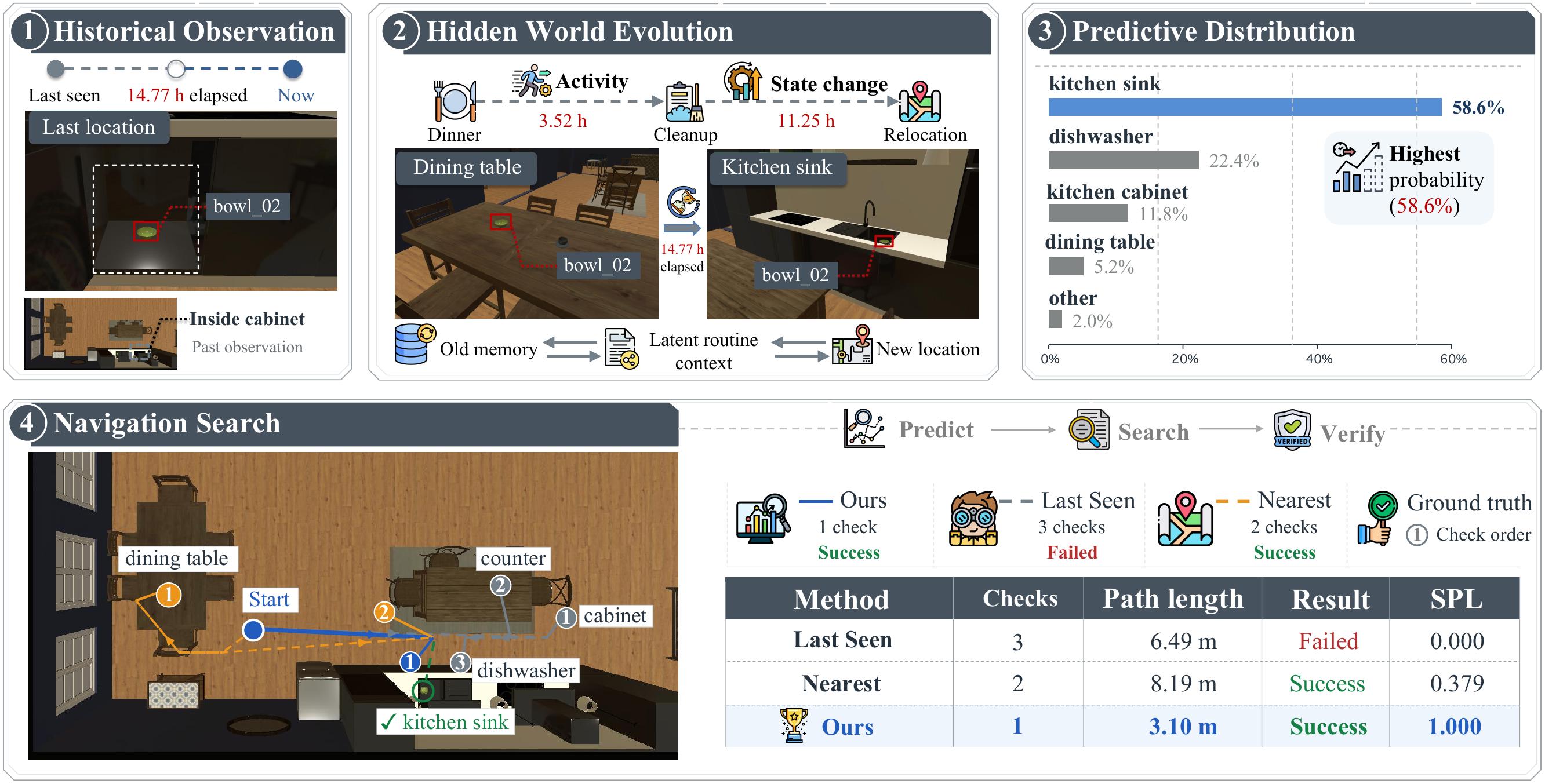}
    \caption{Predictive, last-seen, and nearest-first search on \evolvingworldbench{}.}
    \label{fig:p4d_case_study}
\end{figure}

\paragraph{Online navigation under execution-time dynamics.}
\Cref{fig:p4d_case_study} shows how a routine-conditioned belief avoids an
obsolete last-seen location. To isolate the dynamic closed loop from episodes
in which motion is permitted but no transition is scheduled, \cref{tab:n4-online}
reports the same preselected N4 episodes for every method, with a target
transition scheduled inside a common, fixed post-query window.

\begin{table}[H]
\centering
\caption{N4 online-dynamic navigation with execution-time target motion.}
\label{tab:n4-online}
\scriptsize
\setlength{\tabcolsep}{8pt}
\begin{tabular}{lcccc}
\toprule
\textbf{Method} & \textbf{Dynamic SR $\uparrow$} &
\makecell{\textbf{Online Recovery}\\\textbf{SR $\uparrow$}} &
\makecell{\textbf{Excess Distance}\\\textbf{(m) $\downarrow$}} &
\makecell{\textbf{Revisit Success}\\\textbf{$\uparrow$}}\\
\midrule
SLaTe-PRO \citep{patel2023routine} & 34.7 & 21.5 & 18.9 & 8.7\\
SGM+NEP \citep{kurenkov2023dynamic} & 41.3 & 27.8 & 15.6 & 12.4\\
FlowMaps \citep{argenziano2026flowmaps} & 29.6 & 18.9 & 22.4 & 6.5\\
PredictiveGraphs \citep{saavedraruiz2026predictivegraphs} & 46.8 & 36.7 & 13.2 & 18.9\\
\rowcolor{tblours}\textbf{\evolvingworldnav{} (ours)} &
\textbf{65.2} & \textbf{58.7} & \textbf{8.4} & \textbf{32.6}\\
\bottomrule
\end{tabular}
\end{table}

\Needspace{5\baselineskip}
Relative to PredictiveGraphs, \evolvingworldnav{} improves Dynamic SR by 18.4
points and Online Recovery SR by 22.0 points, while reducing excess distance by
4.8\,m and increasing successful revisits by 13.7 points. These results
support the execution-time contribution: current-time propagation and time-valid
evidence rounds let the agent recover after a target moves, while candidate
reopening turns revisits into useful actions rather than permanent exclusions.

\paragraph{Temporal structure.}
Under paired controls, the SR gain over Last Seen is 22.36 points in routine
worlds but 6.93 points under random transitions (\cref{tab:worlds}).
The agent trails Last Seen in static worlds and Direct Transformer under random
transitions, locating its strongest advantage in learnable temporal regularity.
Across five frozen VLMs, paired Search SR gains over Last Seen range from 14.91
to 25.93 points (20.90 on average; \cref{tab:foundation_models}). Matched
tools, prompts, episodes, and budgets support transfer across controllers.

\noindent\begin{minipage}[t]{0.34\linewidth}
\vspace{0pt}
\paragraph{Generalization and physical deployment.}
Across five benchmark tasks, \evolvingworldnav{} leads on seven of ten metrics (\cref{tab:domains}).
Across 64 matched M20 trials, the agent improves both success and
efficiency (\cref{tab:real-main}); details are in \appref{app:real-world-eval}.
\end{minipage}\hfill
\begin{minipage}[t]{0.63\linewidth}
\vspace{0pt}
\captionof{table}{Primary LYNX M20 search results.}
\label{tab:real-main}
\centering
\scriptsize
\setlength{\tabcolsep}{2.2pt}
\resizebox{\linewidth}{!}{%
\begin{tabular}{@{}lrrrrrr@{}}
\toprule
\textbf{Method} &
\makecell{\textbf{First-Inspection}\\\textbf{SR $\uparrow$}} &
\makecell{\textbf{Search}\\\textbf{SR $\uparrow$}} &
\makecell{\textbf{Recovery}\\\textbf{SR $\uparrow$}} &
\makecell{\textbf{Dist.}\\\textbf{(m) $\downarrow$}} &
\makecell{\textbf{Time}\\\textbf{(s) $\downarrow$}} &
\makecell{\textbf{Inspect.}\\\textbf{$\downarrow$}} \\
\midrule
Last Seen + Search & 17.2 & 26.6 & 12.8 & 56.2 & 284 & 2.19 \\
Time Frequency & 21.9 & 31.3 & 14.0 & 52.9 & 277 & 2.05 \\
Retrieval + Reasoning & 26.6 & 37.5 & 17.1 & 50.3 & 281 & 2.03 \\
\rowcolor{tblours}\textbf{\evolvingworldnav{}} &
\textbf{34.4} & \textbf{48.4} & \textbf{24.3} &
\textbf{43.8} & \textbf{248} & \textbf{1.84} \\
\bottomrule
\end{tabular}%
}
\end{minipage}\par\medskip

Across these evaluations, the gains appear at three levels: cross-benchmark
results improve initial destination selection, N4 isolates recovery under
execution-time motion, and robot trials translate the same belief interface
into shorter paths and fewer inspections. Together, these results suggest that the gains extend beyond
benchmark-specific exploration or controller choice.

\subsection{Ablation Studies}
\label{sec:ablations}

\paragraph{Core components.}
\begin{figure}[!t]
    \centering
    \includegraphics[width=\columnwidth]{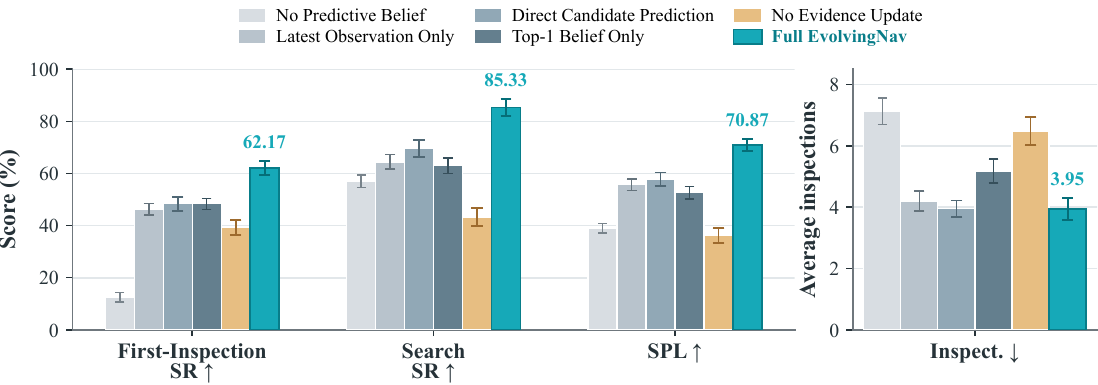}
    \caption{Core component ablation; error bars show five-seed standard deviations.}
    \label{fig:module_ablation}
\end{figure}

The full agent achieves First-Inspection SR of 62.17\%, Search SR of 85.33\%, and
SPL of 70.87\%, with 3.95 inspections (\cref{fig:module_ablation}). Without predictive
belief, First-Inspection SR falls to 12.51\%; Latest Observation Only reduces
Search SR by 20.91 points, indicating that persistent history informs both
initial destination selection and recovery. Direct candidate prediction reduces
First-Inspection/Search SR by 13.84/15.75 points, supporting the persistence--relocation
factorization. Top-1-only search reduces Search SR by 22.41 points, while removing
evidence updates reduces it by 42.02 points and increases inspections to 6.48.
These results support retaining alternative hypotheses and revising them online
to improve search within a fixed action budget.
The pattern separates the roles of the components: predictive belief chooses
where inspection begins, the full distribution preserves fallback hypotheses,
and evidence updating converts new views into route changes. Because No
Evidence Update uses the same predictor, its gap isolates the action-time
filter rather than improved offline state classification.

\begin{wraptable}{r}{0.63\linewidth}
\vspace{-0.7\baselineskip}
\caption{Negative-evidence update ablation.}
\label{tab:evidence-update-ablation}
\centering
\scriptsize
\begin{tabularx}{\linewidth}{@{}l*{4}{>{\centering\arraybackslash}X}@{}}
\toprule
\textbf{Update rule} & \textbf{Rank $\downarrow$} & \textbf{Top-1 $\uparrow$} &
\textbf{NLL $\downarrow$} & \textbf{ECE $\downarrow$}\\
\midrule
No Update & 5.82 & 62.98 & 2.838 & 0.172\\
Hard Removal & 10.09 & 42.69 & 11.514 & 0.204\\
Bayesian Update & 5.25 & 64.71 & \textbf{1.497} & 0.151\\
\rowcolor{tblours}\textbf{Bayesian + Calibration} &
\textbf{4.96} & \textbf{66.93} & 1.530 & \textbf{0.130}\\
\bottomrule
\end{tabularx}
\end{wraptable}

\paragraph{Evidence update and calibration.}
Hard Removal degrades all four metrics (\cref{tab:evidence-update-ablation};
Top-1 in percent). Our calibrated update achieves the best rank, Top-1, and
ECE, although uncalibrated Bayesian updating has slightly lower NLL. These
results support soft, visibility-conditioned evidence over irreversible
removal after a missed detection.

\begin{wraptable}{r}{0.63\linewidth}
\vspace{-0.7\baselineskip}
\caption{Current-state prediction on temporal and LOSO splits.}
\label{tab:predictors}
\centering
\scriptsize
\setlength{\tabcolsep}{2.3pt}
\resizebox{\linewidth}{!}{%
\begin{tabular}{lrrrrrr}
\toprule
& \multicolumn{3}{c}{\textbf{Temporal}} & \multicolumn{3}{c}{\textbf{LOSO}}\\
\cmidrule(lr){2-4}\cmidrule(lr){5-7}
\textbf{Predictor} & \textbf{Top-1 $\uparrow$} & \textbf{MRR $\uparrow$} & \textbf{NLL $\downarrow$}
& \textbf{Top-1 $\uparrow$} & \textbf{MRR $\uparrow$} & \textbf{NLL $\downarrow$}\\
\midrule
Last Seen & 0.00 & 0.1699 & 3.4011 & 50.00 & 0.5850 & 1.8825\\
Time Frequency & 33.85 & 0.5462 & 1.8400 & 49.53 & 0.6166 & 1.6455\\
Direct Transformer & 34.16 & 0.5586 & 1.7292 & 51.71 & 0.6884 & 1.3157\\
Shared semantic prior & 36.02 & 0.5957 & 1.5766 & 59.47 & 0.7462 & 1.0448\\
\rowcolor{tblours} & \textbf{38.24} & \textbf{0.5971} & \textbf{1.4925}
& \textbf{61.18} & \textbf{0.7523} & \textbf{0.9778}\\
\rowcolor{tblours}\multirow{-2}{*}{\textbf{Ours}}
& $\pm1.42$ & $\pm0.0291$ & $\pm0.0867$
& $\pm1.53$ & $\pm0.0176$ & $\pm0.0228$\\
\bottomrule
\end{tabular}%
}
\end{wraptable}

\paragraph{Prediction quality and transfer.}
Relative to the Direct Transformer, our predictor improves Top-1 by 4.08 points
on the temporal split and 9.47 points on LOSO; NLL decreases by 0.2367 and
0.3379, respectively. It also exceeds the semantic
prior, supporting temporal structure beyond location frequency. The larger LOSO
gain supports transfer beyond scene-specific destination statistics, while the
lower NLL indicates improved probabilistic prediction. Additional ablations
are in \appref{app:experimental-results}.

\section{Conclusion}

We study persistent navigation under unobserved changes before and during
execution. \evolvingworldnav{} couples a persistence--relocation belief from 4D
histories with an event-driven predict--observe--replan filter. Across
\evolvingworldbench{}, external benchmarks, and robot trials, it improves initial
inspection and recovery, especially under learnable temporal structure. Future
work will address interacting objects and continuously changing goals. More
broadly, the results suggest that persistent embodied memory should be evaluated
not only by what it stores, but by whether it supports calibrated inference and
evidence-seeking action when the remembered world is no longer current.

\beginappendix
\crefalias{section}{appendix}
\crefalias{subsection}{appendix}
\crefalias{subsubsection}{appendix}
\crefname{appendix}{Appendix}{Appendices}
\Crefname{appendix}{Appendix}{Appendices}

\section{Additional Method Details}
\label{app:method-details}

This section presents the episode interface, memory representation, baseline
specifications, and implementation details supporting the main method.

\subsection{Formal Episode Interface}
\label{app:episode}

Each navigation episode references one time-indexed prediction query. Public fields include schema version, episode and query identifiers, split, task type, scene and household identifiers, world variant, target description, query time, agent start pose, candidate-state identifiers, budgets, and success criteria. Evaluation-private fields include the current state, target pose, valid goal viewpoints, hidden transition class, oracle path, mobility label, and query-time activity. Train and validation releases may expose analysis labels in metadata; the formal test release replaces them with null values and computes grouped metrics in a private evaluator.

The high-level track exposes
\begin{center}
\texttt{NAVIGATE\_TO(state)}, \texttt{INSPECT(state)}, \texttt{EXPLORE},
\texttt{STOP}, \texttt{NOT\_FOUND},
\end{center}
with a fixed navigation backend. The low-level track exposes Habitat-style forward, turn, look, and stop actions. The main paper uses the high-level track to isolate predictive state selection; low-level results measure sensitivity to control and perception.

The observation history available at query time $t_q$ is
\begin{equation}
\mathcal H_{\leq t_q}=\{(I_k,D_k,\xi_k,t_k)\mid t_k\leq t_q\},
\end{equation}
where $I_k,D_k,\xi_k$ denote RGB, depth, and camera pose. The task objective
in \cref{eq:objective} balances successful target verification against travel
and inspection costs. N1--N3 fix the target after the query to isolate
query-time inference and search, whereas N4 allows additional hidden
transitions while the agent moves.

\subsection{Detailed 4D Memory Construction}
\label{app:memory-details}

\begin{figure}[!t]
    \centering
    \includegraphics[width=\linewidth]{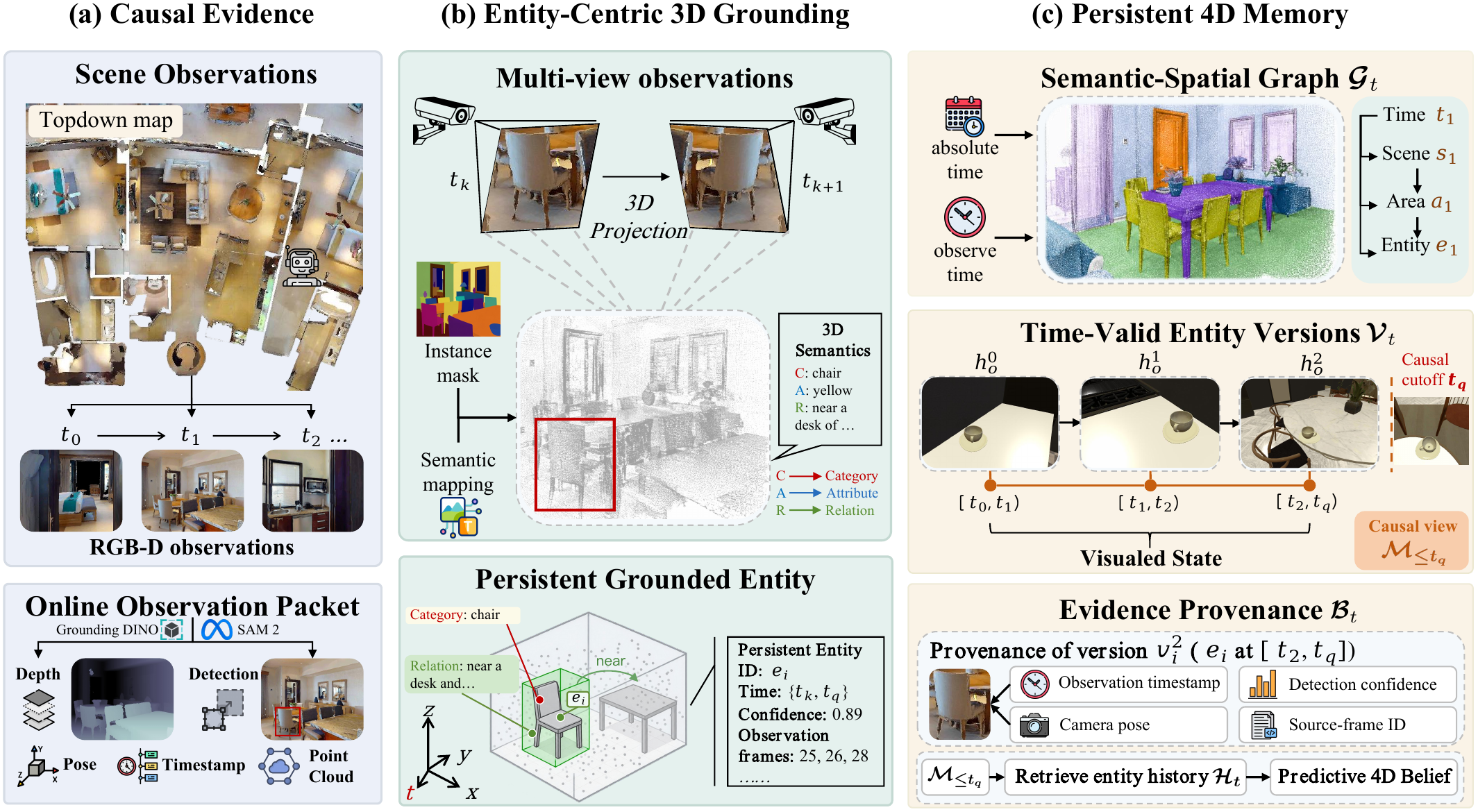}
    \caption{Temporally versioned 4D memory construction.}
    \label{fig:memory}
\end{figure}

\Cref{fig:memory} illustrates how observations are converted into persistent
entity versions. Each observation contributes spatial evidence and temporal
validity, allowing the memory to preserve both previous and current states.
Memory is materialized causally from versioned deltas as defined in
\cref{eq:memory}, retaining semantic--spatial relations, time-valid entity
versions, and observation provenance.
Persistent identity is separated from mutable state: an observed change appends a
time-valid version without erasing earlier states, and each memory query
uses only evidence available at its decision time.

For image coordinate
$\widetilde{\mathbf u}=[u,v,1]^\top$ with depth $D_t(u,v)$, the corresponding
point in world coordinates is
\begin{equation}
\mathbf x^w=\operatorname{proj}_3\!\left(
\mathbf T_{w\leftarrow c,t}
\begin{bmatrix}D_t(u,v)\mathbf K^{-1}\widetilde{\mathbf u}\\1\end{bmatrix}
\right),
\label{eq:backprojection}
\end{equation}
where $\mathbf K$ is the camera intrinsic matrix and $\mathbf T_{w\leftarrow c,t}$ is the camera-to-world transform. Cross-view association uses semantic compatibility, 3D overlap, temporal consistency, and source confidence. The $m$-th version of entity $o$ is
\begin{equation}
h_o^m=([t_{o,s}^m,t_{o,e}^m),s_o^m,c_o^m,f_o^m,\mathcal P_o^m),
\end{equation}
containing a validity interval, candidate state, confidence, visual feature, and evidence handles. An observed state change appends a version without erasing the earlier trajectory, preserving both time-ordered history and observation provenance.

\subsection{Belief Encoding and Training}
\label{app:belief-encoding}

The retrieved history and event representation are defined in
\cref{eq:history,eq:eventtoken}; the Transformer in \cref{eq:history-encoder}
produces the query representation used by both prediction heads. Event tokens
retain entity, state, evidence type, confidence, and visibility alongside
elapsed-time and calendar features. The query token specifies the target,
query time, and last positive state. The persistence probability in
\cref{eq:persistence} measures agreement with that last positive observation,
including departures followed by returns.
The shared pointer head in \cref{eq:pointer} scores every alternative,
including $\mathrm{unknown}$, from the query and candidate representations.
The candidate encoder combines context, functional role, geometry, and instance
identity. The persistence and pointer heads jointly define the normalized
belief in \cref{eq:prior}, trained with current-state negative log-likelihood
without a separate change label or oracle mobility type.

\subsection{Evidence Updates and Online Prediction}
\label{app:belief-control}

\paragraph{Transition-kernel parameterization and training.}
The transition operator in \cref{eq:transition-kernel} is not inferred from a
single marginal belief. It is an additional row-conditional prediction head
that shares the continuous-time history encoder and candidate embeddings with
the query-time predictor. The training set contains chronological tuples
$(H_{\leq t_a},s_{t_a},\Delta t,s_{t_a+\Delta t})$ extracted only from training
world histories. For each tuple, the known source state selects one row and the
future state supervises
\begin{equation}
\mathcal L_{\mathrm{trans}}=-\log
[K_\theta(\Delta t\mid H_{\leq t_a})]_{s_{t_a},s_{t_a+\Delta t}},\qquad
\mathcal L=\mathcal L_{\mathrm{state}}+\lambda_K\mathcal L_{\mathrm{trans}}.
\label{eq:transition-loss}
\end{equation}
Prediction horizons are sampled from the empirical range of action-chunk durations
and candidate arrival times. The softmax over destination states makes every row sum to one. No
transition label, future state, or mobility tag is available at test time. In
N1--N3, where the target is fixed after the query, $K$ is the identity; learned
propagation is activated only for N4.

\paragraph{Decision epochs and ETA.}
The filter begins with $b_0^-=b_q^-$ and $b_0^+=b_0^-$ when no query-time
measurement is available. An epoch is triggered after a short action chunk,
when a view adds sufficient candidate coverage, when the route or ETA changes
materially, when target evidence is obtained, or when the robot reaches an
inspection viewpoint. After a chunk of observed duration
$\Delta t_j=t_{j+1}-t_j$, \cref{eq:filter-predict} first produces the prior at
the new current time. For candidate $i$, the navigation backend then estimates
\begin{equation}
\widehat\tau_{j,i}=\frac{d_{\mathrm{geo}}(x_j,\mathcal W_i)}{\widehat v_j}
+\tau_{\mathrm{inspect}}(i),\qquad
\widehat v_j=\alpha\widehat v_{j-1}+(1-\alpha)\frac{\Delta d_j}{\Delta t_j}.
\label{eq:eta}
\end{equation}
All ETAs are recomputed after obstruction or replanning. Candidate-specific
arrival beliefs forecast future states from the current posterior. They do not
replace the current-time filtering state $b_j^+$ unless the corresponding
prediction horizon has elapsed.

\paragraph{Complete event-driven filter.}
The implementation follows the same ordering at every decision epoch:
\begin{enumerate}[leftmargin=1.55em,itemsep=0pt,topsep=2pt]
    \item Initialize $t_0=t_q$, $b_0^-=b_q^-$, incorporate any valid
    query-time measurement once to obtain $b_0^+$, and initialize an evidence
    ledger for every candidate.
    \item From $b_j^+$, forecast each $b_{j,i}^{\mathrm{arr}}$ to its own ETA,
    score known candidates and \texttt{EXPLORE}, and select the highest-utility
    action.
    \item Execute only a short action chunk and record its actual elapsed time,
    displacement, pose, and RGB-D evidence.
    \item Propagate to the new current time:
    $b_{j+1}^-=b_j^+K_\theta^{(j)}(t_{j+1}-t_j)$.
    \item Create measurements only for candidates whose evidence ledger identifies
    new information; apply \cref{eq:bayes} once per evidence identifier to
    obtain $b_{j+1}^+$. Stop on a verified positive detection.
    \item Update speed and ETAs, forecast all candidate-specific arrival
    beliefs from $b_{j+1}^+$, reopen eligible states, and replan if the preferred
    action changes; otherwise execute the next short chunk.
\end{enumerate}
This explicitly separates the current-time filter
$b_j^-\rightarrow b_j^+$ from the counterfactual arrival forecasts
$\{b_{j,i}^{\mathrm{arr}}\}_i$.

\paragraph{Time-valid evidence rounds without leakage.}
Candidate $i$ maintains an evidence-round index $m(i)$ and
$C_j^{(m)}(i)$, the union of surface samples observed within that round. A
no-detection event is created only if the previously unused coverage satisfies
$\Delta C_j^{(m)}(i)>\delta_C$, with $\delta_C=0.05$. Candidate
surface samples from 4D memory are projected into the live camera, and online
depth marks samples as visible when they lie in the frustum and agree with the
measured depth. The resulting online-estimated coverage and other view features
form $\mathbf f_{j,i}$ for the calibrated detector model $g_\phi$ in
\cref{eq:negative}. Crucially, $\widehat r_{j,i}$ is evaluated on the newly
admitted surface samples and rays, not on the entire overlapping image. One
view can update several candidates with different strengths. Opportunistic
views and deliberate inspections use the same rule, although the latter usually
provide more coverage. A candidate is counted as a sufficiently covered
inspection when $C_j^{(m)}(i)\geq\tau_{\mathrm{cov}}=0.70$; this designation
does not permanently remove it from the candidate set.

Every measurement has a unique evidence identifier and is incorporated once.
Its RGB-D frame and pose remain in memory as provenance, but its likelihood is
not multiplied again or reintroduced as an independent history token into the
current filter. Ground-truth target masks, poses, unoccluded fractions, and
simulator visibility flags are retained only by the private evaluator; an
oracle-visibility diagnostic is reported separately from fair comparisons.
Incremental masking and evidence identifiers do not assert that video frames
are statistically independent; they prevent direct reuse of the same surface
evidence. The conditional-measurement model approximates the remaining dependence,
with detection probabilities calibrated on validation observations. We never
multiply repeated whole-view detection probabilities from overlapping frames.

\paragraph{Reopening candidates and executing unknown.}
There is no permanent exclusion set. Each candidate stores cumulative coverage,
the time of the latest clear negative observation, and cumulative detection probability.
After propagation, a previously inspected candidate becomes eligible when its
belief exceeds $\tau_b=0.10$, predicted return probability
\begin{equation}
P_{\mathrm{return}}(i)=\sum_{r\neq i}b_j^+(r)
K_\theta^{(j)}(r,i;\Delta t)
\label{eq:return-prob}
\end{equation}
exceeds $\tau_{\mathrm{return}}=0.05$, or a new viewpoint offers
$\Delta C(i)>\delta_C=0.05$. Thus a weak or occluded view supports multi-view
reinspection, and a location verified to be empty can regain probability mass after sufficient
world evolution.

Reopening by a new viewpoint alone retains the current round, so only newly
observed geometry contributes evidence. Reopening caused by propagated belief or predicted
return after elapsed time starts round $m(i)+1$ and resets only the coverage gate for that round; the earlier coverage map and negative observations remain in
the provenance ledger. Consequently, the same viewpoint can provide a new
measurement after the world may have changed, while adjacent frames within one
state-validity interval cannot repeatedly suppress the same hypothesis.

The $\mathrm{unknown}$ state invokes an executable \texttt{EXPLORE} action with
\begin{equation}
U_j(\mathrm{EXPLORE})=
\frac{b_j(\mathrm{unknown})\widehat\eta_j^{\mathrm{discover}}}
{c_j^{\mathrm{explore}}+\lambda_{\mathrm{scan}}}.
\label{eq:unknown-action}
\end{equation}
The controller selects a semantic frontier, uncovered region, or unopened
container, performs an open-vocabulary scan, adds discovered states to
$\mathcal C_o$, redistributes unknown mass, and replans. \texttt{NOT\_FOUND} is
allowed only after the exploration budget is exhausted (or no valid frontier
remains) and both unknown mass and remaining searchable mass fall below fixed
validation thresholds. The same filter supports N1--N4 without a
static/dynamic policy router.

\subsection{Full Baseline Specification}
\label{app:baselines}

Last Seen assigns all mass to the latest positive state. Frequency Prior estimates $P(s\mid o)$ from training data. Markov Transition estimates $P(s_{k+1}\mid s_k,o)$; Time-Conditioned Prior additionally conditions on public hour and weekday bins. Instance Hotspot uses only the training trajectory of the target instance. GRU Direct and Transformer Direct receive the same event tokens and candidate encoder as \pfdnav{} but directly normalize candidate logits without the persistence factorization. P4D-Belief Prior uses \cref{eq:prior} once at the start of each episode. Full \evolvingworldnav{} applies \cref{eq:bayes,eq:arrival-belief,eq:planner} at event-driven decision epochs.

The Oracle Activity diagnostic may use the hidden activity and source location but is excluded from fair rankings. Oracle Current State receives the private target state and measures remaining perception and navigation error. External predictive methods are adapted only through their public inputs; any use of privileged simulator state must be identified and excluded from the main comparison.

\subsection{Implementation Details}
\label{app:implementation}

\paragraph{Prediction model and optimization.}
The history encoder uses three Transformer layers, hidden width 128, four
attention heads, maximum history length $K=64$, and dropout 0.1. Candidate
encoders are shared across scenes. The transition scorer $f_\theta$ is a
two-layer MLP of width 128 with GELU activation. The history and candidate
encoders are shared; the persistence, relocation, and transition heads have
separate parameters. We set $\lambda_K=1$ in \cref{eq:transition-loss}. We optimize the model with AdamW using a learning rate of
$3\times10^{-4}$, a weight decay of $10^{-2}$, and a batch size of 64 for at most 100
epochs. Gradients are clipped to an $\ell_2$ norm of 1.0. Early stopping uses
a patience of 10 epochs based on validation NLL, and the checkpoint with the lowest validation
NLL is retained. We use random seeds $\{0,1,2,3,4\}$. Experiments use Habitat-Sim
0.3.3 and Habitat-Lab 0.3.3 on $3\times$ NVIDIA GeForce RTX 5080 GPUs.

\paragraph{Perception and controller.}
The frozen perception stack uses Grounding DINO
\citep{liu2024grounding} with a box threshold of 0.35 and a text threshold of 0.25,
followed by SAM 2 \citep{ravi2025sam} only for mask refinement. The main
high-level agent uses GPT-5.6-Luna as a frozen zero-shot VLM controller, without
navigation-task fine-tuning. Candidate viewpoints and low-level planning are
fixed in the high-level track; a Habitat low-level action track evaluates the
complete embodied stack.

\paragraph{Detection calibration and information boundaries.}
The function $g_\phi$ in \cref{eq:negative} is a lightweight logistic
calibrator that takes as inputs online-estimated candidate coverage, range, viewing
angle, projected size, image quality, category, and validation-set detector
recall. It is fitted only on validation observations and remains frozen
throughout test evaluation. The agent never accesses ground-truth target masks,
ground-truth target poses, unoccluded target fractions, or simulator visibility
flags.

All architecture choices, optimization hyperparameters, calibration models,
perception thresholds, policy thresholds, and checkpoint-selection rules were
determined exclusively from the training and validation splits. The test split
was not used for model selection or hyperparameter tuning and was accessed only
for final evaluation.

\subsection{Experimental Details}
\label{app:experimental-setup}

\paragraph{Benchmark protocols.}
We evaluate the complete agent on \evolvingworldbench{}, FindingDory, and GOAT-Bench
under their native task definitions. FindingDory reports high-level goal
selection and conditional low-level execution; GOAT-Bench reports official SR
and SPL, with Repeat-SR used only as a marked memory-reuse diagnostic. The
cross-domain matrix uses R2R-CE on MP3D, HM3D-OVON on HM3D, native PointNav on
Habitat-GS, SAGE-Bench on InteriorGS, and \evolvingworldbench{} on HSSD. Scores are never
pooled across these heterogeneous tasks, and published values are transferred
only when split, sensors, actions, and success rules match.

\paragraph{Result provenance and comparison scope.}
The entries in \cref{tab:crossbench,tab:domains} come from two explicitly
separated sources. A cell marked \textsuperscript{PR} is transcribed from the
cited paper only when the benchmark version, evaluation split, metric, and
success rule match; the marker applies to that cell rather than to an entire
method row. Every unmarked baseline entry is reproduced by us. Missing values
are left unreported rather than inferred from another task or checkpoint.
Accordingly, \cref{tab:crossbench} evaluates memory and prediction under the task definition of each
benchmark, whereas \cref{tab:domains} measures
cross-environment system performance under native navigation tasks. Neither
table is interpreted as a single architecture-controlled ablation; causal
attribution to our predictive belief is instead provided by the paired-world
control and component ablations.

\paragraph{Test-time information boundary.}
All reproduced methods receive only information public in the target
benchmark. On \evolvingworldbench{}, observations are truncated at the current decision
time. Count and Markov baselines use only the target category, public time, and
latest supported state; structured memories such as HOV-SG and DynaMem replay
the same causal RGB-D and poses; predictive models receive public state or edge
histories and the legal candidate graph. Released navigation agents consume
only their native RGB-D/video window, goal specification, and proprioception.
The paper-reported FindingDory cell specifically uses the
Qwen2.5-VL-3B checkpoint from the cited release and native 96-frame history; it is not recomputed with
our planner. No reproduced method receives the private current state, future
observations, query-time activity, target pose, oracle viewpoint, or simulator
visibility flags. Oracle rows, where present, are diagnostics and are excluded
from fair rankings.

\paragraph{Training and model selection.}
Non-parametric baselines are estimated from the public training split, with
smoothing and time-bin settings selected on the validation split. Internal neural baselines use the
same event tokens, candidate encoder, training scenes, and validation rule as
\pfdnav{}. External predictive architectures that require target-domain fitting
are trained only on \evolvingworldbench{} days 0--79 and selected on days 80--84; days
85--89 remain test-only. When an official checkpoint exists for a native
benchmark, we retain that checkpoint and its prescribed observation window.
Task-adapted systems are labeled separately from zero-shot systems in
\cref{tab:domains}; interface conversion is not counted as navigation-policy
training.

\paragraph{Prediction splits.}
Each scene history follows a 90-day chronological split: days 0--79 are used
for training, days 80--84 for validation, and days 85--89 for testing. Temporal
evaluation pools the held-out test periods and reports 319 changed-only queries;
leave-one-scene-out evaluation rotates the held-out scene while preserving the
same chronological boundaries and reports 644 change-balanced queries. All
predictors share query identities, time-ordered histories, and legal candidate
sets. The main table reports five-seed variation; no future observation or
private transition label is available at inference time.

\paragraph{Real-world protocol.}
LYNX M20 is the primary quantitative platform, with X30 and Lite3 used for
transfer. The primary M20 comparison uses 64 matched trials, whereas X30 and
Lite3 use matched 32-block transfer subsets. All methods receive the same
pre-query histories, candidate locations, start conditions, perception interface,
and robot-specific low-level stack. Success requires autonomous online
confirmation within three inspections during a 360-second episode. Missed
detections, navigation failures, timeouts, and human interventions count as
failures. Indoor/outdoor balance, platform specifications, and condition-wise
breakdowns are provided in \appref{app:real-world}.

\paragraph{Reporting controls.}
We use two method-preserving adaptation regimes. Memory-only and
prediction-only systems retain their native representation or predictor, map
their outputs onto the legal candidate states, and use the shared planner,
inspection viewpoints, and action budget; this isolates temporal reasoning
from low-level control. End-to-end navigation agents retain their released
perception, mapping, and policy, with adapters limited to sensor conventions,
task-goal formatting, action vocabulary, and \texttt{STOP} semantics. We do not
add our VLM planner or predictive belief to those agents. The cross-domain
scores therefore measure full-system compatibility, while the controlled
\evolvingworldbench{} and real-robot comparisons support module-level claims. Results are
stratified by world variant, staleness, transition class, mobility regime, and
generalization split. Paired routine/static/random episodes retain scene,
query, and marginal placement factors while changing only the hidden
transition mechanism.

The \evolvingworldbench{} entries in
\cref{tab:crossbench,tab:domains,tab:foundation_models,fig:module_ablation,tab:temporal-memory-ablation}
use the same N1--N5 test episodes, protocol, action budget, and metric
definitions. Full-agent results are obtained from separate stochastic runs on these
episodes; small numerical differences therefore do not indicate a
change in the test set. Controlled gains are computed within each matched
comparison. Routine-world and N4-transition diagnostics condition on specified
subsets and are identified separately.

\section{Benchmark Construction and Evaluation Protocol}
\label{app:benchmark-protocol}

This section documents episode validity, leakage controls, mobility assignment,
and the metrics used for all reported comparisons.

\subsection{Construction and Audit Details}
\label{app:benchmark-construction}

\paragraph{Design principles.}
\evolvingworldbench{} follows four principles: temporal continuity, past-only observability,
physical executability, and controlled dynamics. Examples are sampled from
persistent scene histories rather than independent placements. Public memories
contain only evidence available to the robot by the query time. Candidate states
must admit collision-free placements and reachable, visibility-checked inspection
viewpoints. Paired worlds hold the scene, target, query, and marginal placement
factors fixed while varying the hidden transition mechanism. Together, these
controls separate predictive reasoning from scene frequency, route geometry,
and privileged simulator access.

\paragraph{Source alignment and accounting.}
Source records constrain the generator rather than serving as episode templates.
We distinguish raw source rows or sequences ($N_{\rm raw}$), records after
normalization specific to each source ($N_{\rm std}$), distinct normalized records
referenced by at least one retained generation rule ($N_{\rm used}$), and total
rule references with reuse allowed ($N_{\rm ref}$). Table~\ref{tab:source-accounting}
reports counts from the audited consumption manifests; reuse therefore does not
inflate the number of independent source observations.

\begin{wraptable}{r}{0.45\linewidth}
\vspace{-0.8\baselineskip}
\caption{Audited source contributions.}
\label{tab:source-accounting}
\centering
\scriptsize
\setlength{\tabcolsep}{1.25pt}
\begin{tabular}{lrrrr}
\toprule
\textbf{Source} & $\mathbf{N_{\rm raw}}$ & $\mathbf{N_{\rm std}}$ & $\mathbf{N_{\rm used}}$ & $\mathbf{N_{\rm ref}}$ \\
\midrule
CASAS Aruba & 1,602,820 & 20,252 & 428 & 480 \\
ARAS & 5,184,000 & 5,080 & 312 & 360 \\
HD-EPIC & 59,454 & 10,786 & 742 & 864 \\
ParaHome & 212 seq. & 860 & 186 & 216 \\
OPPORTUNITY & 869,387 & 2,551 & 318 & 372 \\
HOMER+ (sim.) & 65 seq. & 1,991 & 356 & 420 \\
\bottomrule
\end{tabular}
\end{wraptable}

The raw record unit is a sensor row for CASAS, ARAS, and OPPORTUNITY, an annotation row
for HD-EPIC, and a sequence for ParaHome and HOMER+. CASAS events are mapped to
the room ontology and split into sessions at inactivity gaps exceeding 900 seconds;
the resulting occupancy, time-of-day, and weekday statistics parameterize
household schedules. ARAS and OPPORTUNITY sensor streams are segmented into
activity intervals and normalized to the shared activity, room, and object
vocabularies, supplying complementary activity compatibility and temporal-context
statistics. HD-EPIC narration verbs and nouns are mapped through canonical
action and object aliases to estimate object--action frequencies, without
assigning HSSD destinations. ParaHome annotations and object transforms provide
displacement and transition-timing evidence. HOMER+ remains explicitly marked
as simulated and contributes only long-horizon activity order and terminal-state
priors, not independent human observations or test-time model inputs.

\paragraph{Mapping rules and measured parameters.}
All adapters emit a common schema containing time, activity, actor context,
object category, source state, destination state, and provenance. Measured
quantities comprise CASAS occupancy timing, ARAS/OPPORTUNITY activity context,
HD-EPIC action--object frequencies, ParaHome displacement and transition timing,
and HOMER+ simulated order priors. Legal receptacles, reachable
viewpoints, and collision-free placements are derived from HSSD geometry.
Category-to-receptacle allowlists, affordance constraints, stochastic exception
rates, paired static/routine/random interventions, the 90-day horizon, and query
sampling are benchmark-authored and labeled as such. Thus no source trajectory
is copied verbatim and no source count is expanded into a claimed number of
independent human observations. The final manifest spans 54 scenes and 4,860
scene-days and contains approximately 239.19k dynamic state changes and 803.68k
task instances after physical-validity, observability, and leakage audits.

\paragraph{Household generation and behavioral checks.}
For each HSSD scene \citep{khanna2024hssd}, resident schedules, room preferences,
orderliness, object ownership, and placement habits are fixed at the household
level. Activities, resident region trajectories, and object transitions are
generated jointly in chronological order over multiple days; stochastic
events, delayed returns, and exceptions prevent deterministic evolution.
Household profiles, transition rules, and task templates instantiate executable
N1--N5, VLN, and EQA episodes. Every relocation is attributed to an activity,
resident transition, object lifecycle event, or tidying event. Personal objects
alternate between placed and carried states with owner-specific return habits;
irregular objects use affordance-valid destinations without temporal or
actor-specific predictability. Audits check that personal objects do not simply
track their owners and that actor or activity identity does not predict irregular
destinations beyond affordance. The audited configuration yields 42.6 region
transitions per resident-day, 6.56 relocations per personal object-day, 52.6\%
within-region relocations, and zero terminal collisions.

\begin{figure}[!t]
    \centering
    \includegraphics[width=0.94\linewidth]{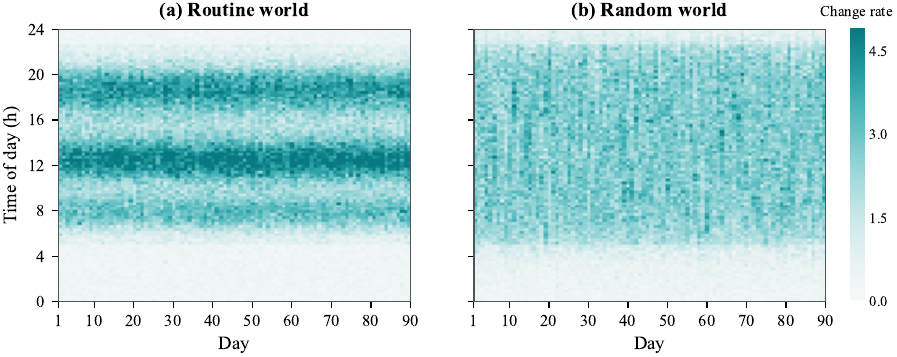}
    \caption{Temporal change density in paired routine and random worlds.}
    \label{fig:routine-random-heatmaps}
\end{figure}

The realized transition density in \cref{fig:routine-random-heatmaps} provides a
direct check of the paired-world intervention. Routine worlds retain repeatable,
activity-linked time bands, whereas random worlds spread changes across the day
while matching the marginal movement process. The control therefore removes
predictable temporal structure without changing the 90-day horizon or scene setup.
The ``Change rate'' color scale is the mean number of object relocations per
scene-day in each one-hour time-of-day bin.

\subsection{Automated Audit and Human Spot Checks}
\label{app:human-review}

All generated episodes undergo automated quality checks before retention. The
checks verify entity identity and event ordering, legal and collision-free
placements, reachable candidate viewpoints, consistency between world states
and rendered observations, and the query-time cutoff that prevents future
information from entering public histories. They also verify target--episode
correspondence, valid not-found tracks, and task budgets that permit executable
search. Failed episodes are regenerated or excluded; the audit manifest records
the check outcomes for every retained episode.

Human assessment uses stratified spot checks rather than reviewing every
episode. Five trained researchers sample across data sources, scenes, mobility
regimes, and task types. Using a structured checklist, they assess the
plausibility of relocation reasons and household routines, the semantic
consistency of negative evidence, and the clarity and relevance of language
goals. A supervising researcher adjudicates disputed or high-risk sampled
cases. Reviewers record accept, revise, or reject decisions for the sampled
items; systematic issues prompt corrections to the generation rules and
renewed automated checks of the affected episodes. The human review log
identifies the sampled items and decisions, while the automated manifest
accounts for the full retained set.

\paragraph{Candidate metadata and query sampling.}
Chronological Habitat replay records navigability, geodesic cost, camera frustum
coverage, unoccluded target fraction, and success viewpoints for each candidate.
These fields support the physical validity and visibility checks described in
\appref{app:audit}; candidates without collision-free placements and reachable
inspection viewpoints are excluded. Query times are sampled from both the natural temporal
distribution and windows relevant to the mechanisms under study, including routine stages and
intervals after a personal object is dropped. Each navigation query is augmented
with a start pose, candidate inspection viewpoints, a path budget, and frozen
success conditions. The patrol policy cannot access current object states,
resident locations, hidden activities, or generator latents. Query-time activity
and hidden transitions are retained only by the evaluator.

\paragraph{Language tasks and release documentation.}
EQA questions cover existence, location, temporal order, activity association,
count, comparison, and compositional relations over the same time-indexed
memories.
Answers are evaluated against a normalized ontology, and questions requiring
future evidence are excluded. VLN replaces the object label with a language goal
whose referents and constraints are grounded in the same versioned scene history.
The broader task annotations support studies beyond the prediction and navigation
experiments emphasized in this paper. Across tracks, only memory representations,
current-state beliefs, and resulting decisions differ between controlled methods.
Dataset cards and audit summaries document source mappings, authored assumptions,
physical validity, candidate coverage, and leakage tests.

\paragraph{Splits and leakage checks.}
Each 90-day history uses days 0--79 for training, days 80--84 for validation,
and days 85--89 for testing. Complete temporal sequences remain within the same split so
that no decision uses observations from a later time. Held-out homes are reserved for
cross-scene evaluation; paired static/routine/random members remain in the same
split. Near-duplicate queries and transition groups are assigned to splits as indivisible units.
Serialized releases are checked for future timestamps, private event fields,
answer-correlated filenames or indices, repeated event groups across splits,
inconsistent paired variants, invalid physics, unreachable goals, and accidental
patrol dependence on hidden state. Episode-level checks and mobility assignment
are further specified in \appref{app:audit,app:mobility}.

\subsection{Mobility Regimes and Task Protocols}

Mobility labels are used only for stratified evaluation and are hidden from the
agent.

\begin{table}[!htbp]
\caption{Mobility regimes in \evolvingworldbench{}.}
\label{tab:mobility-regimes}
\centering
\scriptsize
\setlength{\tabcolsep}{3.5pt}
\begin{tabularx}{\linewidth}{lXXX}
\toprule
\textbf{Regime} & \textbf{Mechanism} & \textbf{Frozen signal} & \textbf{Role}\\
\midrule
Stable/rare & Infrequent events & Home/return tendency & Persistence control\\
Routine & Activity state machine & Chain-specific locations & Routine prediction\\
Personal & Placed/carried lifecycle & Owner placement habit & Long-term memory\\
Irregular & Affordance sampling & Affordance only & Uncertainty control\\
\bottomrule
\end{tabularx}
\end{table}

\begin{table}[!htbp]
\caption{Prediction and navigation protocols in \evolvingworldbench{}.}
\label{tab:tasks}
\centering
\scriptsize
\setlength{\tabcolsep}{3.5pt}
\begin{tabularx}{\linewidth}{cXXX}
\toprule
& \textbf{Task} & \textbf{Decision signal} & \textbf{Primary measure}\\
\midrule
P   & Current-state prediction & Region/receptacle belief & R@$k$ / NLL\\
N1  & Predictive navigation & Event-filtered belief & First-Inspection SR\\
N2  & Belief-guided search & Full belief + cost & Search SR / cost\\
N3  & Evidence-aware replanning & Updated posterior & Recovery SR\\
N4  & Online-dynamic navigation & Arrival-time belief & Online Recovery SR\\
N5  & Cross-scene generalization & Any protocol above & Held-out SR\\
\bottomrule
\end{tabularx}
\end{table}
\FloatBarrier

\subsection{Episode Validity and Leakage Audit}
\label{app:audit}

An episode is retained only if: (i) the target has at least one positive
pre-query observation; (ii) the target state is known to the private evaluator
and physically instantiated without collision; (iii) at least one valid
viewpoint lies on the NavMesh; (iv) the start is on the NavMesh, at least
$3$ meters from the target, and does not reveal it; and (v) all public histories
terminate at or before the query. A stop is successful only when the robot is
within $1$ meter geodesic distance of a valid viewpoint, the private evaluator
measures target visible fraction
$v_{\mathrm{target}}^{\mathrm{GT}}\geq\tau_{\mathrm{success}}=0.20$, and the
agent identifies the correct target instance or category. This threshold is
used only by the private evaluator and is never exposed to the agent.

The three coverage quantities have distinct roles:
$\Delta C>0.05$ triggers a new negative-evidence update,
$C\geq0.70$ defines a sufficiently covered inspection, and
$v_{\mathrm{target}}^{\mathrm{GT}}\geq0.20$ defines evaluation success only.
The first two are estimated online from candidate geometry and depth; the third
is private evaluator state.

We audit serialized examples for future timestamps, hidden event fields, filenames or indices correlated with answers, duplicate event groups across splits, and inconsistent paired variants. Matched static, routine, and random variants must have identical public query fields. Invalid simulator episodes are reported separately and are not counted as ordinary failures.

\subsection{Mobility-Type Assignment}
\label{app:mobility}

The frozen rule is
\begin{equation}
\operatorname{type}(o)=
\begin{cases}
\mathrm{stable}, & r_{\mathrm{move}}(o)<\tau_{\mathrm{move}},\\
\mathrm{activity}, & G_{\mathrm{act}}(o)\geq\tau_{\mathrm{act}},\\
\mathrm{personal}, & C_{\mathrm{habit}}(o)\geq\tau_{\mathrm{habit}},\\
\mathrm{irregular}, & \text{otherwise}.
\end{cases}
\end{equation}
All thresholds are selected on training data before navigation evaluation. When the generator specifies a mobility profile, the same statistics are used to verify that realized trajectories exhibit the intended behavior. Every group must cover several categories, instances, and scenes, and no category may uniquely reveal one mobility type.

\subsection{Metric Definitions}
\label{app:metrics}

Let $S_i$ be episode success and $L_i$ the executed path. For episodes whose
target remains fixed after the query, $L_{i,\mathrm{fix}}^*$ is the shortest
feasible path to a valid target viewpoint from the same start. For an online
dynamic episode, $L_{i,\mathrm{dyn}}^*$ is the minimum travel distance along
a time-feasible trajectory to a viewpoint at which the target can be verified
in its scheduled state, under the same start, NavMesh, and action/time budget. The
dynamic oracle knows the private transition schedule for evaluation only; the
agent does not. Writing $L_i^{\mathrm{ref}}$ for the applicable fixed or dynamic
oracle path, we use
\begin{equation}
\mathrm{SPL}=\frac{1}{N}\sum_{i=1}^{N}S_i
\frac{L_i^{\mathrm{ref}}}{\max(L_i,L_i^{\mathrm{ref}})}.
\end{equation}
The N1--N5 aggregate includes N4 and applies this reference episode by episode;
the fixed-target and online-dynamic slices use their respective reference
paths when computing SPL. The static oracle is never applied to a moving target.
First-Inspection SR records whether the first candidate to receive sufficient
cumulative coverage contains the target; opportunistic evidence and replanning
before that inspection are part of the policy. A sufficiently covered en-route
view counts as the first inspection itself. Recovery SR conditions on an
unsuccessful first inspection and measures whether the target is subsequently
found. Excess path uses $L_i-L_i^{\mathrm{ref}}$, with the dynamic oracle for
moving-target episodes. Before any N4 rollout, each episode receives a common
post-query evaluation window $[t_q,t_q+T_i]$ and a fixed schedule of hidden
target transitions. The dynamic subset contains exactly the episode identifiers
with a scheduled change of target state inside this window, independent of the
destinations, trajectories, or termination times of individual methods. All methods are scored
on this same set. The complementary no-transition slice is kept separate.
Dynamic SR is the success rate on the preselected dynamic set. Online Recovery SR uses
the shared subset whose predeclared transition invalidates the
pre-transition target state; only successful verification after that transition counts
as recovery, and an early stop does not change the denominator. Revisit Success
is the percentage of dynamically reopened candidate visits that recover the
target.

We compute calibration metrics on the full candidate belief before navigation and after each update. ECE groups predictions by maximum confidence, the Brier score evaluates all candidates, and true-state rank captures changes beyond Top-1 accuracy. Search-cost curves plot success against the inspection budget and distance traveled, providing comparisons across budgets.

\section{Additional Experimental Results}
\label{app:experimental-results}

This section provides extended quantitative results omitted from the main paper
because of space constraints.

\subsection{Cross-Environment Navigation}

SR and SPL are reported in percent for all environments.
\Cref{tab:domains} compares five native benchmark tasks:
\evolvingworldnav{} leads on seven of ten metrics, including SR and SPL on
HM3D-OVON, PointNav, and \evolvingworldbench{}. Uni-LaViRA leads on R2R-CE SR,
while NaVILA leads on R2R-CE and SAGE-Bench SPL. These task-specific exceptions
limit any claim of uniform superiority across navigation protocols.

\begin{table}[!t]
\caption{Cross-environment navigation.}
\label{tab:domains}
\centering
\scriptsize
\setlength{\tabcolsep}{3.6pt}
\resizebox{\textwidth}{!}{%
\begin{tabular}{lcccccccccc}
\toprule
& \multicolumn{4}{c}{\hdr{\textbf{Standard Simulation Worlds}}} &
\multicolumn{4}{c}{\hdr{\textbf{Photorealistic Simulation}}} &
\multicolumn{2}{c}{\hdr{\textbf{Evolving Simulation Worlds}}}\\
\cmidrule(lr){2-5}\cmidrule(lr){6-9}\cmidrule(lr){10-11}
& \multicolumn{2}{c}{\textbf{R2R-CE}} &
\multicolumn{2}{c}{\textbf{HM3D-OVON}} &
\multicolumn{2}{c}{\textbf{PointNav}} &
\multicolumn{2}{c}{\textbf{SAGE-Bench}} &
\multicolumn{2}{c}{\textbf{\evolvingworldbench{}}}\\
\cmidrule(lr){2-3}\cmidrule(lr){4-5}\cmidrule(lr){6-7}\cmidrule(lr){8-9}\cmidrule(lr){10-11}
\textbf{Method} & \textbf{SR} & \textbf{SPL} & \textbf{SR} & \textbf{SPL} &
\textbf{SR} & \textbf{SPL} & \textbf{SR} & \textbf{SPL} & \textbf{SR} & \textbf{SPL}\\
\midrule
\rowcolor{tblmemory}\multicolumn{11}{c}{\emph{Navigation-trained or task-adapted}}\\
MTU3D \citep{zhu2025move} & 8.8 & 6.5 & \paperreported{40.8} & \paperreported{12.1} & 75.8 & 72.7 & 25.9 & 15.6 & 35.2 & 30.7\\
NaVid \citep{zhang2024navid} & \paperreported{37.4} & \paperreported{35.9} & 60.3 & 36.0 & 15.3 & 13.4 & 17.4 & 15.1 & 31.0 & 22.6\\
Uni-NaVid \citep{zhang2024uni} & \paperreported{47.0} & \paperreported{42.7} & \paperreported{39.5} & \paperreported{19.8} & 15.9 & 13.2 & 26.4 & 23.1 & 55.3 & 41.7\\
NaVILA \citep{cheng2024navila} & \paperreported{54.0} & \paperreported{49.0} & 18.3 & 9.7 & 78.4 & 77.5 & 39.0 & \textbf{34.0} & 35.2 & 30.7\\
\midrule
\rowcolor{tblnav}\multicolumn{11}{c}{\emph{Zero-shot navigation or task adapters}}\\
HSGM \citep{Li_2026_CVPR} & \paperreported{47.9} & \paperreported{32.8} & 40.0 & 32.5 & 37.2 & 35.3 & 11.8 & 9.4 & 17.0 & 12.5\\
VLFM \citep{yokoyama2024vlfm} & 12.0 & 9.6 & \paperreported{35.2} & \paperreported{19.6} & 53.0 & 42.2 & 24.4 & 18.8 & 35.0 & 25.8\\
AO-Planner \citep{chen2025affordances} & \paperreported{25.5} & \paperreported{16.6} & 30.9 & 22.0 & 76.6 & 74.7 & 27.4 & 19.7 & 37.0 & 29.3\\
TANGO \citep{ziliotto2025tango} & 14.3 & 9.87 & \paperreported{35.5} & \paperreported{19.5} & 35.8 & 33.5 & 22.7 & 14.7 & 31.0 & 23.3\\
Uni-LaViRA \citep{ding2026uni} & \textbf{\paperreported{60.7}} & \textbf{\paperreported{47.7}} & \paperreported{60.0} & \paperreported{40.5} & 81.3 & 77.7 & 35.6 & 28.7 & 65.2 & 39.4\\
MSGNav \citep{huang2026msgnav} & 11.6 & 9.4 & \paperreported{48.3} & \paperreported{27.0} & 37.2 & 34.8 & 23.5 & 17.5 & 26.7 & 19.6\\
\rowcolor{tblours}\textbf{\evolvingworldnav{} (ours)} & 55.8 & 43.1 &
\textbf{64.2} & \textbf{51.7} & \textbf{86.2} & \textbf{79.8} &
\textbf{42.4} & 32.1 & \textbf{86.6} & \textbf{68.2}\\
\bottomrule
\end{tabular}
}
\vspace{1pt}

\parbox{\textwidth}{\scriptsize
\textsuperscript{PR} Paper-reported under the cited native protocol; unmarked
baseline scores are our reproductions under the method-preserving protocol in
\appref{app:experimental-setup}.}
\end{table}

\subsection{Full Prediction Results}
\label{app:full-prediction-results}

The temporal split contains 319 changed-only queries and LOSO contains 644
change-balanced queries. We report the mean $\pm$ standard deviation over five
seeds for our method; Top-1 is a percentage, while MRR and NLL are unscaled.

\begin{table}[H]
\centering
\scriptsize
\setlength{\tabcolsep}{3pt}
\caption{Full current-state prediction results.}
\label{tab:predictors-full}

\resizebox{\linewidth}{!}{%
\begin{tabular}{lrrrrrr}
\toprule
\textbf{Predictor}
& \textbf{Temporal Top-1 $\uparrow$}
& \textbf{MRR $\uparrow$}
& \textbf{NLL $\downarrow$}
& \textbf{LOSO Top-1 $\uparrow$}
& \textbf{MRR $\uparrow$}
& \textbf{NLL $\downarrow$} \\
\midrule

Uniform Random
& 9.32
& 0.2607
& 2.4509
& 11.96
& 0.3050
& 2.2980 \\

Last Seen
& 0.00
& 0.1699
& 3.4011
& 50.00
& 0.5850
& 1.8825 \\

Object Frequency
& 30.75
& 0.5476
& 1.7397
& 50.47
& 0.6441
& 1.8455 \\

Time Frequency
& 33.85
& 0.5462
& 1.8400
& 49.53
& 0.6166
& 1.6455 \\

Markov
& 18.94
& 0.4435
& 2.3772
& 49.38
& 0.6286
& 1.6982 \\

Activity-Markov
& 25.16
& 0.4582
& 2.2436
& 49.84
& 0.6221
& 1.6183 \\

Random Forest
& 20.50
& 0.4851
& 1.9181
& 58.07
& 0.7283
& 1.1441 \\

Packed MLP
& 29.50
& 0.5225
& 2.1036
& 58.07
& 0.7173
& 1.3105 \\

GRU
& 32.61
& 0.5471
& 1.8279
& 54.04
& 0.7016
& 1.2558 \\

Direct Transformer
& 34.16
& 0.5586
& 1.7292
& 51.71
& 0.6884
& 1.3157 \\

Shared semantic prior
& 36.02
& \textbf{0.5957}
& 1.5766
& 59.47
& 0.7462
& 1.0448 \\

\rowcolor{tblours}
& \textbf{38.24}
& \textbf{0.5971}
& \textbf{1.4925}
& \textbf{61.18}
& \textbf{0.7523}
& \textbf{0.9778} \\
\rowcolor{tblours}\multirow{-2}{*}{\textbf{Ours}}
& $\pm 1.42$
& $\pm 0.0291$
& $\pm 0.0867$
& $\pm 1.53$
& $\pm 0.0176$
& $\pm 0.0228$ \\

\bottomrule
\end{tabular}%
}
\end{table}

\subsection{EvoWorld-Bench Detailed Results}
\label{app:moved-tables}

Table~\ref{tab:mainnav} complements the main benchmark table with search cost
and recovery after an unsuccessful first inspection; lower is better for the two
cost columns. Table~\ref{tab:foundation_models} tests whether paired navigation
gains persist across different VLMs under matched tools, prompts, episodes, and
action budgets. In the VLM study, SR is reported in percent and SPL is
unscaled.

\begin{table}[H]
\caption{EvoWorld-Bench routine-world search cost and recovery.}
\label{tab:mainnav}
\centering
\scriptsize
\setlength{\tabcolsep}{8.0pt}
\begin{tabular}{lccc}
\toprule
\textbf{Method} & \textbf{Inspect. $\downarrow$} &
\makecell{\textbf{Distance (m)}\\\textbf{$\downarrow$}} &
\makecell{\textbf{Recovery SR (\%)}\\\textbf{$\uparrow$}}\\
\midrule
\rowcolor{tblnav}\multicolumn{4}{c}{\emph{Navigation and persistent memory}}\\
VLFM \citep{yokoyama2024vlfm} & 4.85 & 18.72 & 22.41\\
LagMemo GLUE \citep{zhou2025lagmemo} & 3.92 & 13.24 & 45.62\\
HOV-SG \citep{werby2024hovsg} & 3.36 & 12.08 & 51.24\\
DynaMem \citep{liu2024dynamem} & 2.51 & 11.59 & 63.93\\
\midrule
\rowcolor{tblpredict}\multicolumn{4}{c}{\emph{Predictive world and state modeling}}\\
SLaTe-PRO \citep{patel2023routine} & 4.40 & 14.58 & 29.00\\
SGM+NEP \citep{kurenkov2023dynamic} & 4.08 & \textbf{7.42} & 27.56\\
FlowMaps \citep{argenziano2026flowmaps} & 4.66 & 16.51 & 20.27\\
PredictiveGraphs \citep{saavedraruiz2026predictivegraphs} & 3.71 & 10.48 & 17.96\\
\midrule
\rowcolor{tblours}\textbf{\evolvingworldnav{} (ours)} & \textbf{2.18} & 8.36 & \textbf{68.42}\\
\bottomrule
\end{tabular}
\end{table}

\evolvingworldnav{} requires the fewest inspections (2.18) and achieves the
highest Recovery SR (68.42\%), improving over DynaMem by 4.49 points after an
unsuccessful first inspection. SGM+NEP travels the shortest average distance
(7.42\,m), while \evolvingworldnav{} remains close at 8.36\,m. Together with
the First-Inspection, Search SR, and SPL results in \cref{tab:crossbench}, these
complementary metrics indicate that retaining uncertainty and updating evidence
reduce redundant inspections and support recovery from an incorrect initial
prediction.

\begin{table}[H]
\caption{VLM robustness on EvoWorld-Bench.}
\label{tab:foundation_models}
\centering
\scriptsize
\setlength{\tabcolsep}{3.5pt}
\begin{tabular}{@{}lccccccc@{}}
\toprule
\multirow{2}{*}{\textbf{Different VLMs}} &
\multicolumn{3}{c}{\textbf{Last-seen Agent}} &
\multicolumn{3}{c}{\textbf{\evolvingworldnav{}}} &
\multirow{2}{*}{\makecell{\textbf{$\Delta$ Search}\\\textbf{SR $\uparrow$}}}\\
\cmidrule(lr){2-4}\cmidrule(lr){5-7}
& \makecell{\textbf{First-Inspection}\\\textbf{SR $\uparrow$}} &
\makecell{\textbf{Search}\\\textbf{SR $\uparrow$}} & \textbf{SPL $\uparrow$} &
\makecell{\textbf{First-Inspection}\\\textbf{SR $\uparrow$}} &
\makecell{\textbf{Search}\\\textbf{SR $\uparrow$}} & \textbf{SPL $\uparrow$} & \\
\midrule
GPT-4o & 31.88 & 48.94 & 0.3215 & \textbf{59.72} & \textbf{71.86} & \textbf{0.6027} & \textbf{+22.92}\\
GPT-5.5 & 42.36 & 51.31 & 0.3732 & \textbf{63.48} & \textbf{73.57} & \textbf{0.6531} & \textbf{+22.26}\\
GPT-5.6-Luna & 46.92 & 57.43 & 0.3970 & \textbf{63.87} & \textbf{83.36} & \textbf{0.6639} & \textbf{+25.93}\\
Qwen2.5-VL-3B & 16.26 & 26.55 & 0.1502 & \textbf{38.27} & \textbf{45.02} & \textbf{0.3727} & \textbf{+18.47}\\
Qwen2.5-VL-32B & 18.84 & 31.30 & 0.1591 & \textbf{42.44} & \textbf{46.21} & \textbf{0.3713} & \textbf{+14.91}\\
\midrule
\multicolumn{7}{r}{\textbf{Average gain}} & \textbf{+20.90}\\
\bottomrule
\end{tabular}
\end{table}

\FloatBarrier
\subsection{Real-World Search Case}

\begin{figure}[H]
    \centering
    \includegraphics[width=\textwidth]{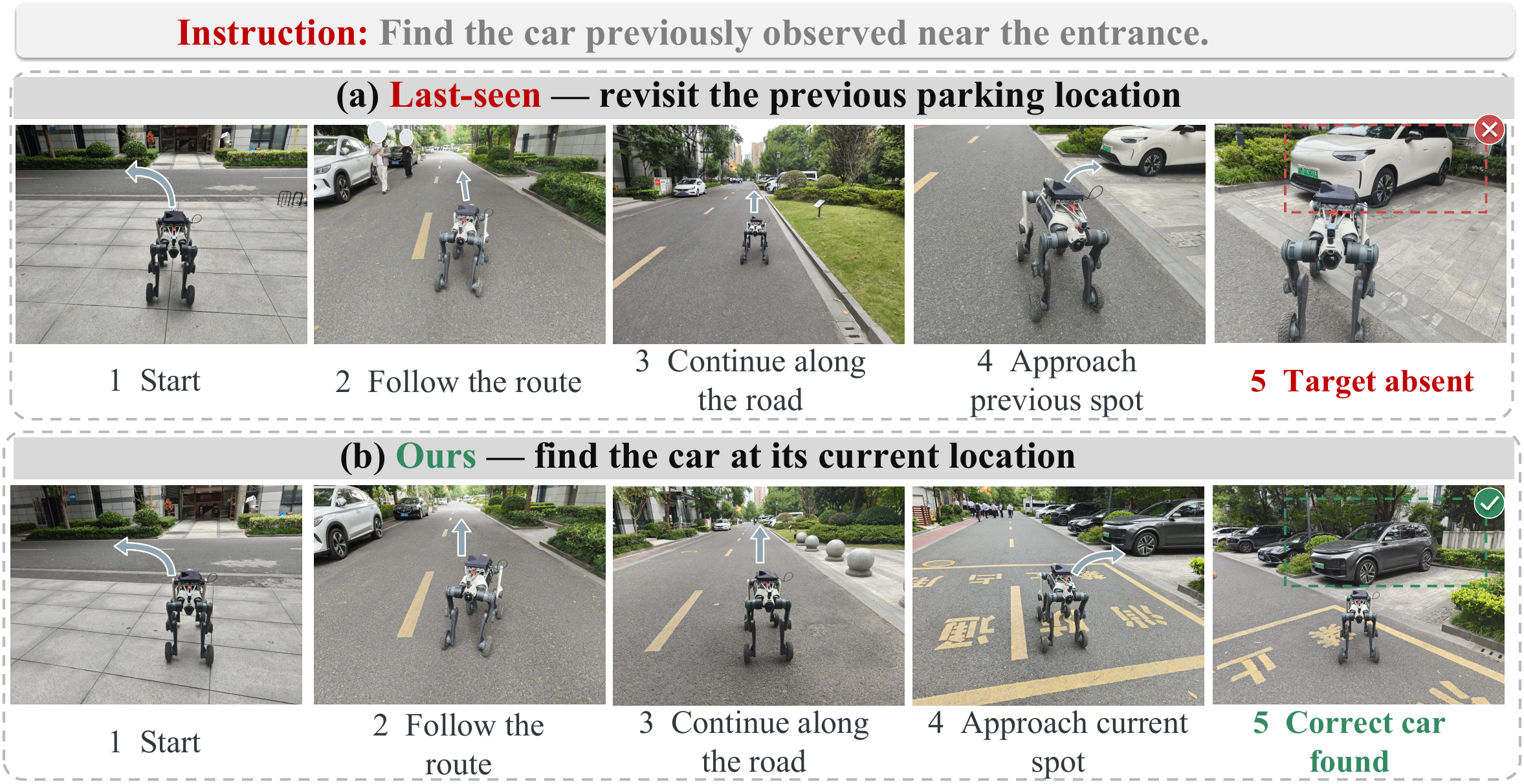}
    \caption{Outdoor car search under stale and predictive memories.}
    \label{fig:real-world-car-case}
\end{figure}

\subsection{Fine-Grained Ablation Analysis}
\label{app:fine-grained-ablation}
\label{app:prediction-evidence}

The main-paper ablation isolates the five decisions that define the complete
agent. Here we examine implementation choices within temporal memory,
evidence updating, and the predictor while preserving the same perception,
candidate set, inspection viewpoints, and low-level navigator.

\subsubsection{Temporal and Memory Ablations}

SR and SPL are reported in percent throughout this analysis.

\begin{table}[!htbp]
\caption{Temporal and memory ablations.}
\label{tab:temporal-memory-ablation}
\centering
\scriptsize
\setlength{\tabcolsep}{3.5pt}
\begin{tabularx}{0.88\linewidth}{@{}l>{\raggedright\arraybackslash}Xcccc@{}}
\toprule
\textbf{Group} & \textbf{Variant} &
\makecell{\textbf{First-Inspection}\\\textbf{SR $\uparrow$}} &
\makecell{\textbf{Search}\\\textbf{SR $\uparrow$}} & \textbf{SPL $\uparrow$} &
\textbf{Inspect. $\downarrow$}\\
\midrule
\multirow{3}{*}{Temporal encoding}
& w/o elapsed-time encoding & 48.37 & 80.28 & 64.82 & 4.36\\
& w/o calendar context & 47.75 & 76.11 & 66.20 & 4.05\\
& w/o both temporal cues & 40.84 & 76.03 & 64.98 & 4.12\\
\midrule
\multirow{2}{*}{Historical evidence}
& w/o instance-specific history & 57.57 & 80.77 & 66.61 & 4.30\\
& w/o negative history & 45.86 & 79.53 & 63.83 & 4.44\\
\midrule
& $K=8$ & 39.55 & 63.77 & 56.00 & 4.57\\
& $K=16$ & 45.88 & 69.48 & 65.77 & 4.37\\
& $K=32$ & 56.25 & 79.60 & 68.40 & 4.15\\
\rowcolor{tblours}\multirow{-4}{*}{History capacity} & \textbf{$K=64$ (Full)} & \textbf{60.23} &
\textbf{84.63} & \textbf{70.34} & \textbf{3.98}\\
\bottomrule
\end{tabularx}
\end{table}
\FloatBarrier

Elapsed-time and calendar ablations distinguish irregular observation gaps from
daily and weekly periodic context. Removing both cues reduces First-Inspection SR by
19.39 points relative to the full model. Instance-specific trajectories provide
smaller but consistent gains, whereas removing negative history reduces
First-Inspection SR by 14.37 points and SPL by 6.51 points. Increasing capacity from $K=8$ to $K=64$
raises First-Inspection SR by 20.68 points, Search SR by 20.86 points, and SPL by 14.34
points while reducing the average inspection count from 4.57 to 3.98. These
trends show that performance benefits from temporally structured evidence over a
sufficiently long history, rather than merely the most recent observations.

\subsubsection{Evidence Update Ablations}

\begin{figure}[!t]
\centering
\includegraphics[width=\linewidth]{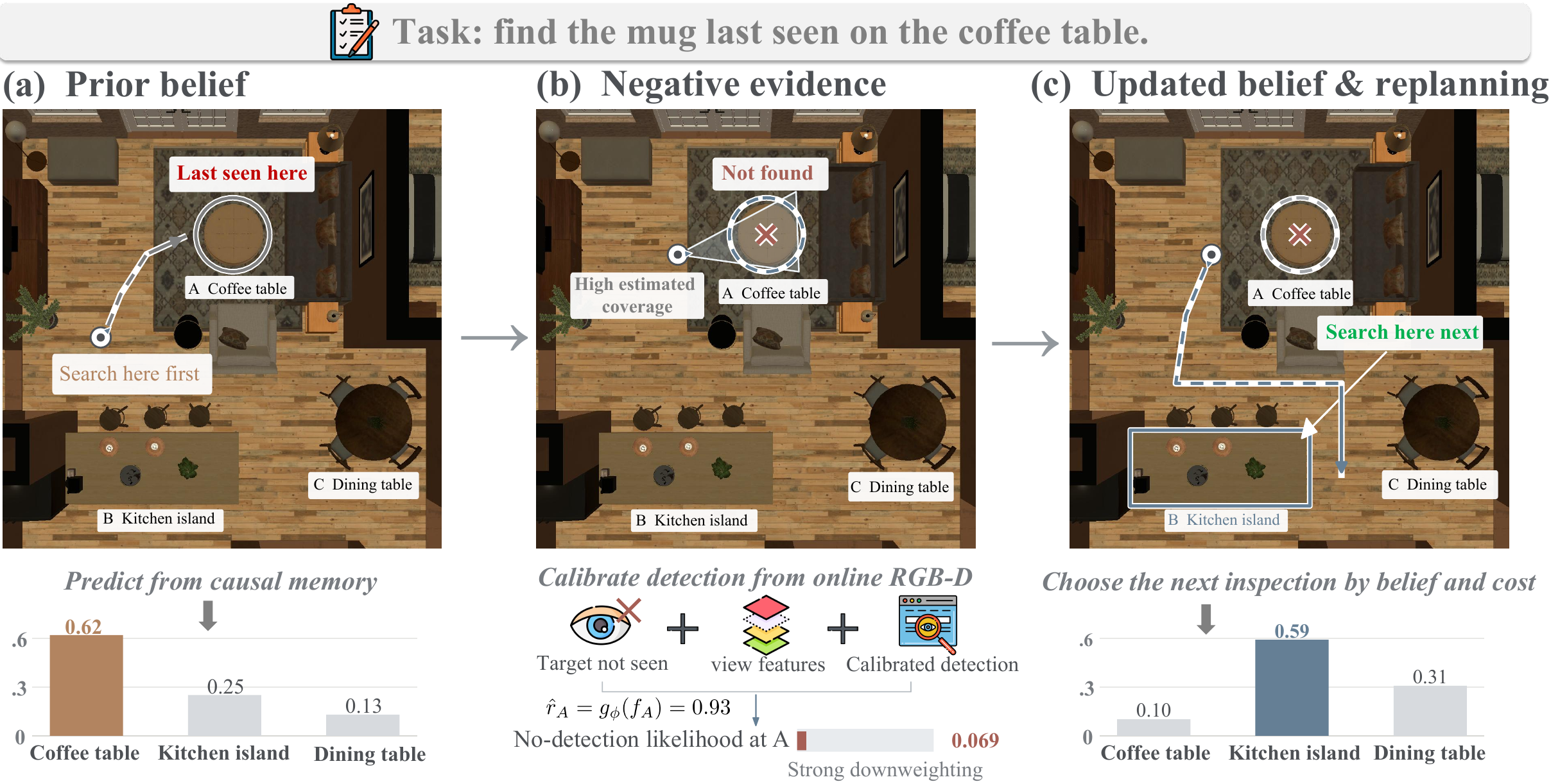}
\caption{Evidence-aware belief updating and replanning after a negative inspection.}
\label{fig:agent-evidence-update}
\end{figure}

\Cref{fig:agent-evidence-update} illustrates one deliberate-inspection event in
the same stepwise filter used for opportunistic views. Causal memory initially
assigns probability 0.62 to the last-seen coffee table. The frozen logistic
calibrator maps the online view features $\mathbf f_{j,i}$ to
$\widehat r_{j,i}=0.931$; when the target is not detected, the likelihood at
that state is therefore 0.069. The posterior consequently shifts from $(0.62,0.25,0.13)$ to
$(0.10,0.59,0.31)$ and the planner selects the kitchen island next. The inspected
state retains nonzero mass, reflecting residual perceptual uncertainty rather
than an irreversible deletion.

\Cref{tab:evidence-update-ablation} in the main paper compares the four update
rules. No Update retains the prior; Hard Removal assigns zero probability after
each unsuccessful inspection; Bayesian Update applies a soft likelihood; and
our calibrated variant conditions that likelihood on online view features.

\subsubsection{Predictor Architecture}

\begin{table}[!htbp]
\caption{Predictor architecture on changed-only queries.}
\label{tab:predictor-architecture-ablation}
\centering
\scriptsize
\setlength{\tabcolsep}{7pt}
\begin{tabular}{lccc}
\toprule
\textbf{Predictor} & \textbf{Top-1 $\uparrow$} & \textbf{MRR $\uparrow$} &
\textbf{NLL $\downarrow$}\\
\midrule
GRU & 32.61 & 0.5471 & 1.8279\\
Direct Transformer & 34.16 & 0.5586 & 1.7292\\
\rowcolor{tblours}\textbf{Structured Transformer (ours)} &
$\mathbf{38.24\pm1.42}$ & $\mathbf{0.5971\pm0.0291}$ &
$\mathbf{1.4925\pm0.0867}$\\
\bottomrule
\end{tabular}
\end{table}

All three predictors use the same observation tokens and candidate encoder.
The Direct Transformer normalizes candidate logits directly, whereas the
structured model separates persistence from relocation. This controlled comparison
therefore isolates architectural factorization from memory content and action
selection.

\subsection{Evolution-Aware Diagnostics}
\label{app:evolution-diagnostics}

\begin{figure}[!b]
\centering
\includegraphics[width=\linewidth]{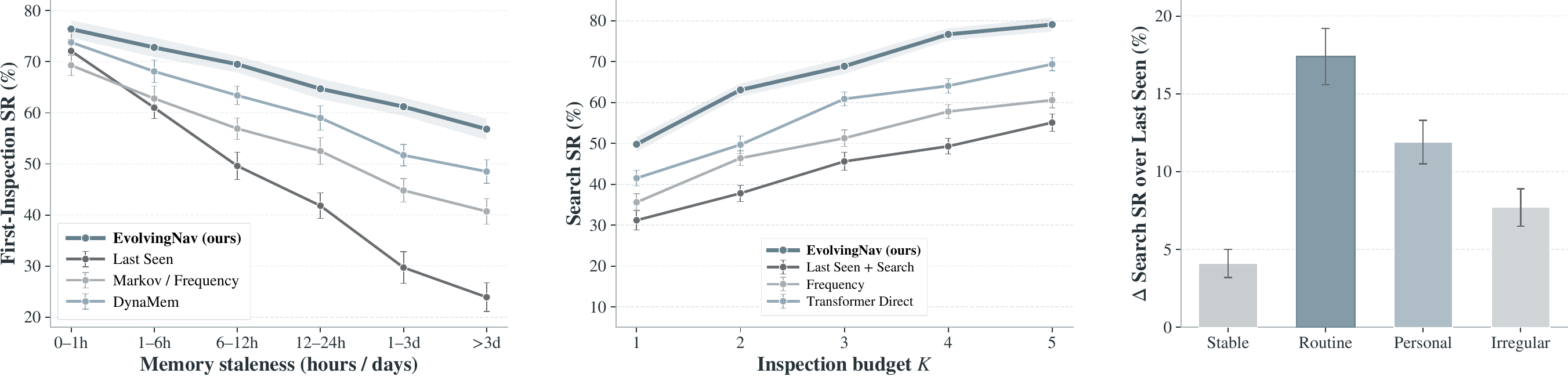}
\caption{Predictive diagnostics under memory staleness and search budgets.}
\label{fig:predictive-memory}
\end{figure}

\paragraph{Paired-world control.}
The paired-world comparison in \cref{tab:worlds} holds scenes and queries fixed
while changing only the world transition mechanism. Last Seen reaches 84.38 SR in static worlds but falls to
38.49 under routine evolution, exposing the failure of stale memory. In contrast,
\evolvingworldnav{} reaches 60.85 routine-world SR, improving by 22.36 points over Last
Seen and by 12.03 points over Transformer Direct. Under matched random motion,
the gain over Last Seen decreases to 6.93 points and \evolvingworldnav{} trails
Transformer Direct by 1.27 points. This contrast localizes its main advantage
to learnable temporal regularity rather than a scene-frequency or generic
search shortcut.

\begin{table}[H]
\caption{Paired-world SR and routine gain (\%).}
\label{tab:worlds}
\centering
\scriptsize
\setlength{\tabcolsep}{2.6pt}
\resizebox{0.85\linewidth}{!}{%
\begin{tabular}{lcccc}
\toprule
\textbf{Method} & \textbf{Static SR} & \textbf{Routine SR} & \textbf{Random SR} &
\makecell{\textbf{Routine gain}\\\textbf{over Last Seen}}\\
\midrule
Last Seen & \textbf{84.38} & 38.49 & 23.19 & --\\
Markov Transition & 80.22 & 32.57 & 20.64 & $-5.92$\\
Direct Transformer & 69.79 & 48.82 & \textbf{31.39} & $+10.33$\\
\rowcolor{tblours}\textbf{\evolvingworldnav{}} & 82.19 & \textbf{60.85} & 30.12 & \textbf{$+22.36$}\\
\bottomrule
\end{tabular}
}
\end{table}

\paragraph{Effect of memory staleness and inspection budget.}
\Cref{fig:predictive-memory} complements these aggregate tables by showing
how performance changes with temporal staleness, inspection budget, and hidden
mobility type. These episode-level diagnostics are not inferred from aggregate
success rates.

\begin{figure*}[!t]
\centering
\includegraphics[width=0.96\textwidth]{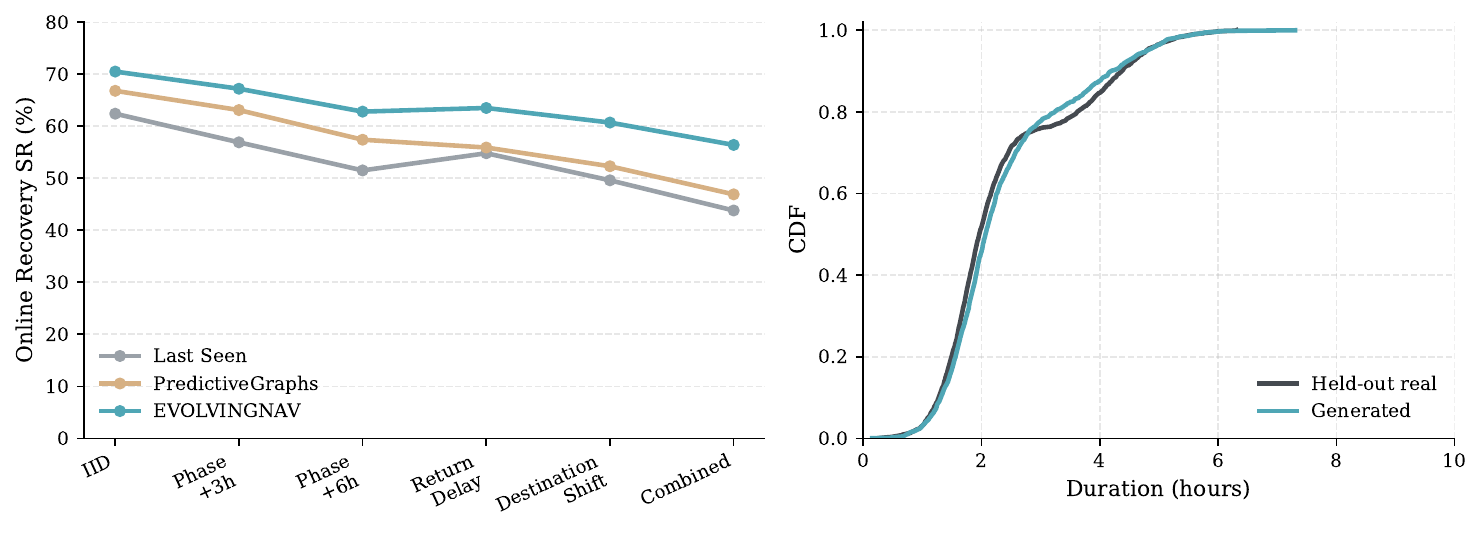}
\caption{Recovery under behavior shift and temporal grounding to held-out traces.}
\label{fig:behavior-shift-grounding}
\end{figure*}

\paragraph{Behavior-shift and trace-grounding diagnostic.}
\Cref{fig:behavior-shift-grounding} tests whether the learned regularities are
limited to the nominal timing of the benchmark generator. Online Recovery SR falls
for all methods under phase offsets, return delays, destination shifts, and
their combination, but \evolvingworldnav{} achieves higher recovery than Last Seen and
PredictiveGraphs in every condition. The duration CDF further shows that the
generated transition durations track the broad temporal scale of held-out real
traces without exactly reproducing them. This diagnostic supports robustness to
moderate behavior shift and temporal grounding beyond a single generator
setting; it is not presented as evidence of unrestricted out-of-distribution
generalization.

\section{Real-World Evaluation}
\label{app:real-world-eval}

This section describes the robot platforms, presents M20 results by environment
and temporal condition, and provides qualitative cases.

\subsection{Robot Platforms}
\label{app:robot-platforms}

We use three commercial DEEP Robotics platforms: LYNX M20, X30, and Lite3.
Platform-specific perception and low-level control use a
common interface, while the memory, prediction, belief update, and high-level
search components remain unchanged.

\begin{table}[H]
\caption{Robot platforms for real-world evaluation.}
\label{tab:robot-platforms}
\centering
\scriptsize
\setlength{\tabcolsep}{8pt}
\begin{tabular}{lllll}
\toprule
\textbf{Platform} & \textbf{Morphology} & \textbf{Standing size} & \textbf{Mass} & \textbf{Endurance / range} \\
\midrule
LYNX M20 & Wheel-legged & $820\times430\times570$ mm & $\sim$35 kg & 3 h / 15 km unloaded \\
X30 & Quadruped & $1000\times695\times470$ mm & 56 kg & 2.5--4 h / $\geq$10 km \\
Lite3 (LiDAR) & Quadruped & $610\times370\times496$ mm & 13.5 kg & 1.5--2 h / 2.7 km \\
\bottomrule
\end{tabular}
\end{table}

Manufacturer-rated endurance and range are descriptive specifications, not
experimental outcomes. Absolute completion time is compared only between
methods executed on the same platform.

\subsection{Environment Breakdown}
\label{app:real-world}

Table~\ref{tab:real-domain} separates indoor and outdoor trials. The outdoor
setting has longer routes and lower success for all methods, while
\evolvingworldnav{} improves performance in both environments.

\begin{table}[H]
\caption{LYNX M20 results by environment.}
\label{tab:real-domain}
\centering
\scriptsize
\setlength{\tabcolsep}{5pt}
\begin{tabular}{lrrrrrr}
\toprule
& \multicolumn{3}{c}{\textbf{Indoor}} & \multicolumn{3}{c}{\textbf{Outdoor}} \\
\cmidrule(lr){2-4}\cmidrule(lr){5-7}
\textbf{Method} & \textbf{First-Inspection} & \textbf{Search SR} & \textbf{Dist.} &
\textbf{First-Inspection} & \textbf{Search SR} & \textbf{Dist.} \\
\midrule
Last Seen + Search & 18.8 & 31.3 & 28.6 & 15.6 & 21.9 & 83.8 \\
Time Frequency & 25.0 & 34.4 & 27.2 & 18.8 & 28.1 & 78.6 \\
Retrieval + Reasoning & 31.3 & 40.6 & 25.1 & 21.9 & 34.4 & 75.5 \\
\textbf{\evolvingworldnav{}} & \textbf{40.6} & \textbf{53.1} & \textbf{22.4}
& \textbf{28.1} & \textbf{43.8} & \textbf{65.2} \\
\bottomrule
\end{tabular}
\end{table}

\begin{figure*}[!t]
    \centering
    \includegraphics[width=0.92\textwidth]{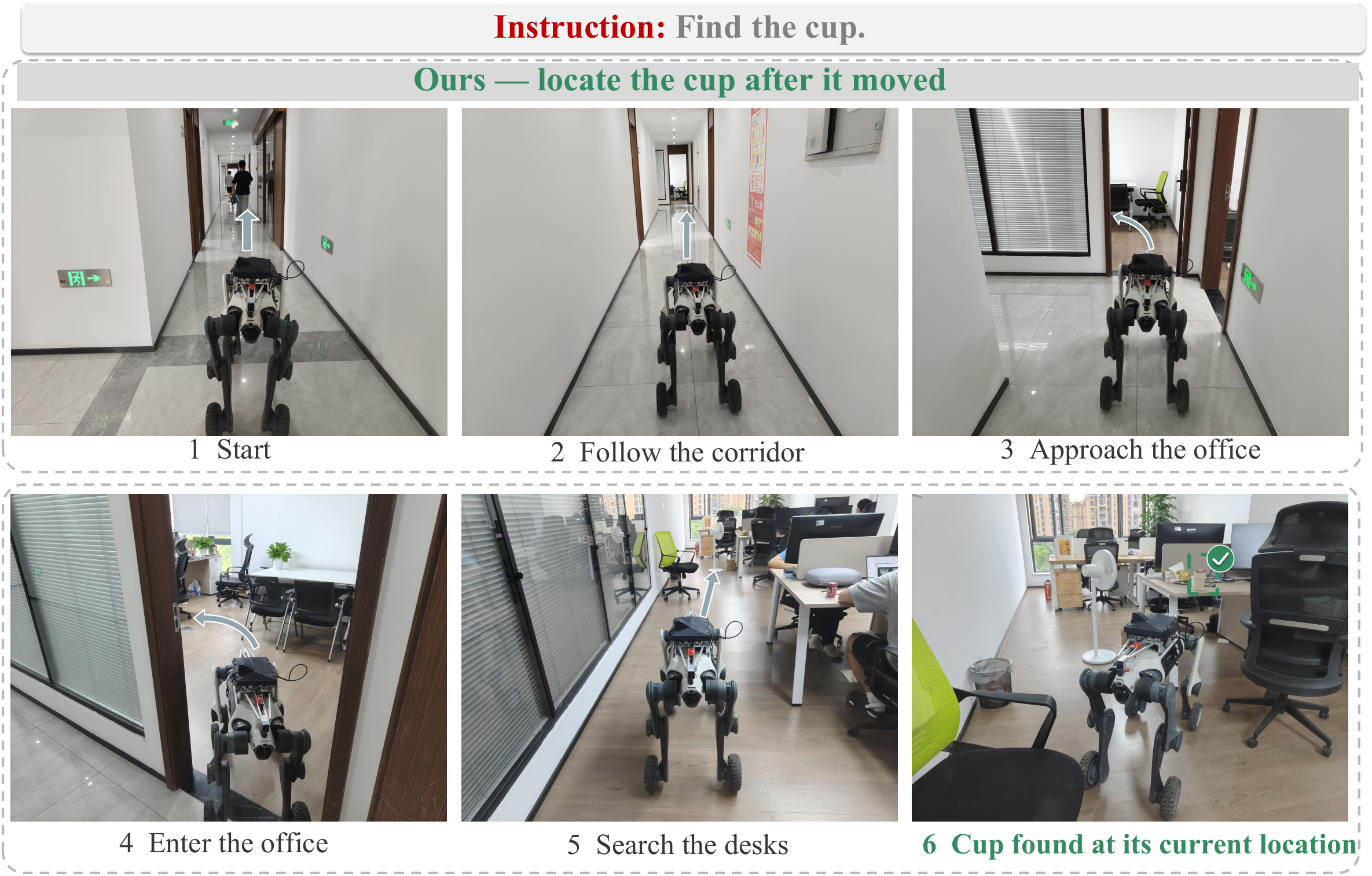}
    \vspace{2pt}
    \includegraphics[width=0.92\textwidth]{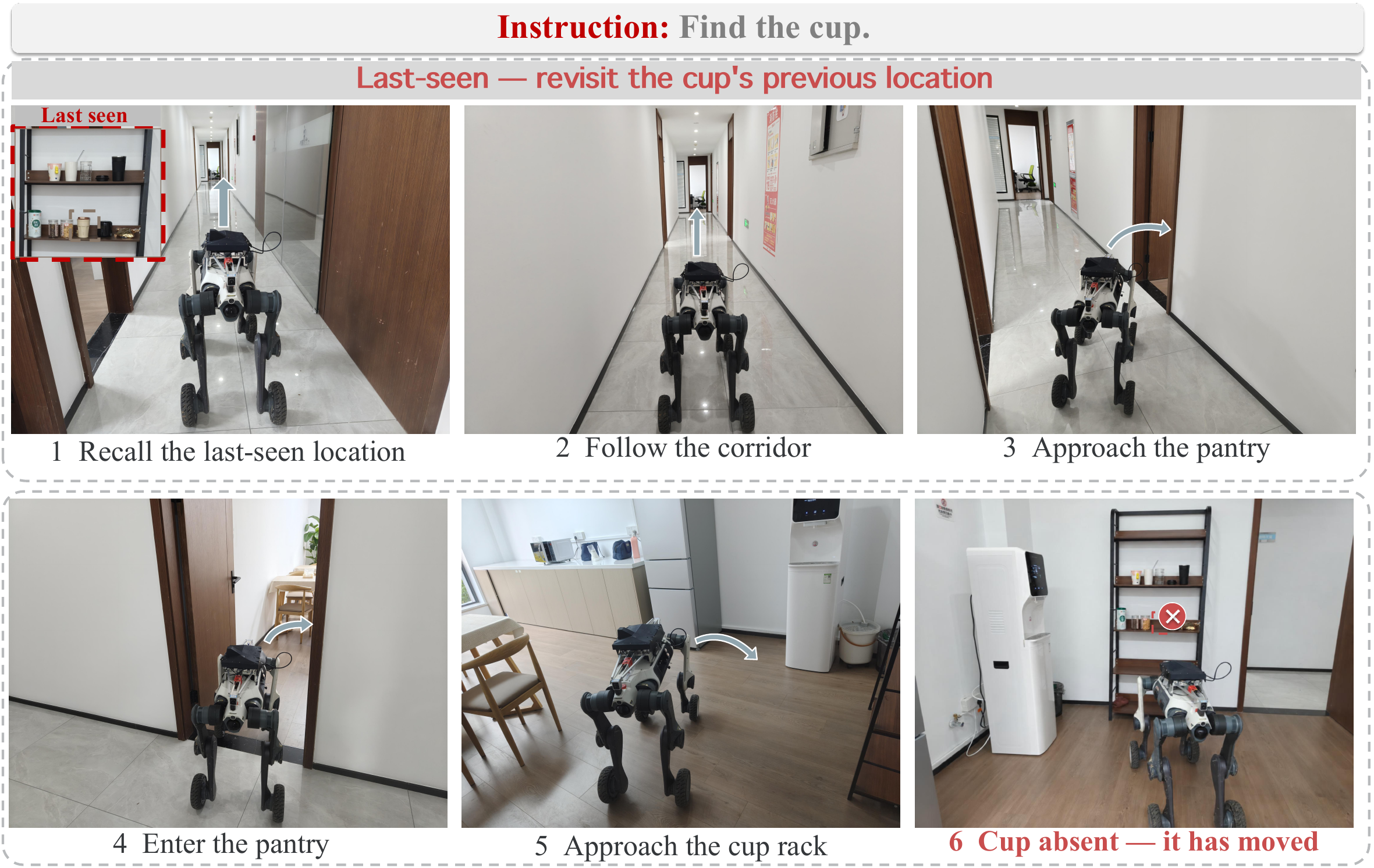}
    \caption{Indoor cup search with predictive belief (top) and stale Last Seen memory (bottom).}
    \label{fig:real-world-cup-comparison}
\end{figure*}

\subsection{Temporal Condition Analysis}
\label{app:temporal-analysis}

\begin{wraptable}{r}{0.60\linewidth}
\vspace{-0.7\baselineskip}
\centering
\caption{LYNX M20 Search SR by temporal condition (\%).}
\label{tab:real-condition}
\scriptsize
\setlength{\tabcolsep}{2.5pt}
\begin{tabular}{@{}lrrrr@{}}
\toprule
\textbf{Method} & \textbf{Unchanged} &
\makecell{\textbf{Routine-}\\\textbf{consistent}} &
\makecell{\textbf{Weak-}\\\textbf{routine}} &
\makecell{\textbf{Broken-}\\\textbf{routine}} \\
\midrule
Last Seen + Search & \textbf{68.8} & 18.8 & 12.5 & 6.3 \\
Time Frequency & 50.0 & 50.0 & 18.8 & 6.3 \\
Retrieval + Reasoning & 62.5 & 43.8 & 25.0 & 18.8 \\
\textbf{\evolvingworldnav{}} & 62.5 & \textbf{68.8} & \textbf{37.5} & \textbf{25.0} \\
\bottomrule
\end{tabular}
\end{wraptable}

\paragraph{Temporal conditions.}
\Cref{tab:real-condition} shows that predictive memory is most useful when
recurring changes provide a usable historical signal. Last Seen remains strong
when nothing changes, whereas all methods degrade as regularity weakens. This
breakdown distinguishes the benefits of routine modeling from improvements
attributable to generic search.

\par\medskip
\Needspace{14\baselineskip}
\begin{wraptable}[10]{r}{0.60\linewidth}
\vspace{-0.7\baselineskip}
\centering
\caption{Cross-platform navigation transfer.}
\label{tab:real-transfer}
\scriptsize
\setlength{\tabcolsep}{2.5pt}
\begin{tabular}{@{}llrrr@{}}
\toprule
\textbf{Platform} & \textbf{Method} & \textbf{First-Inspection SR} &
\textbf{Search SR} & \textbf{Dist. (m)} \\
\midrule
\multirow{2}{*}{LYNX M20} & Last Seen + Search & 18.8 & 25.0 & 55.4 \\
& \textbf{\evolvingworldnav{}} & \textbf{37.5} & \textbf{50.0} & \textbf{44.6} \\
\midrule
\multirow{2}{*}{X30} & Last Seen + Search & 18.8 & 25.0 & 54.1 \\
& \textbf{\evolvingworldnav{}} & \textbf{31.3} & \textbf{43.8} & \textbf{46.8} \\
\midrule
\multirow{2}{*}{Lite3} & Last Seen + Search & 18.8 & 31.3 & 51.8 \\
& \textbf{\evolvingworldnav{}} & \textbf{25.0} & \textbf{43.8} & \textbf{45.9} \\
\bottomrule
\end{tabular}
\end{wraptable}

\paragraph{Recovery and cross-platform transfer.}
Following unsuccessful first inspections, \evolvingworldnav{} recovered 9/37 M20
episodes (24.3\%). \Cref{tab:real-transfer} reports the matched 32-block
transfer comparison for X30 and Lite3 alongside the corresponding M20 subset;
the primary M20 result still uses all 64 trials. This evaluates transfer, not
platform parity.

\subsection{Qualitative Cases}
\label{app:real-world-cases}

The outdoor comparison in \cref{fig:real-world-car-case} isolates stale memory
in a car-search task: Last Seen revisits the previous entrance-side parking
location and observes that the target is absent, whereas \evolvingworldnav{}
reaches its current location. The two-row indoor comparison in
\cref{fig:real-world-cup-comparison} shows the same failure mode for a cup:
\evolvingworldnav{} routes to the predicted office location, while Last Seen
first inspects the now-empty pantry rack. These traces visualize representative
outcomes and do not add observations to the aggregate success-rate measurements.

\FloatBarrier

\begin{figure}[!t]
    \centering
    \includegraphics[width=\textwidth]{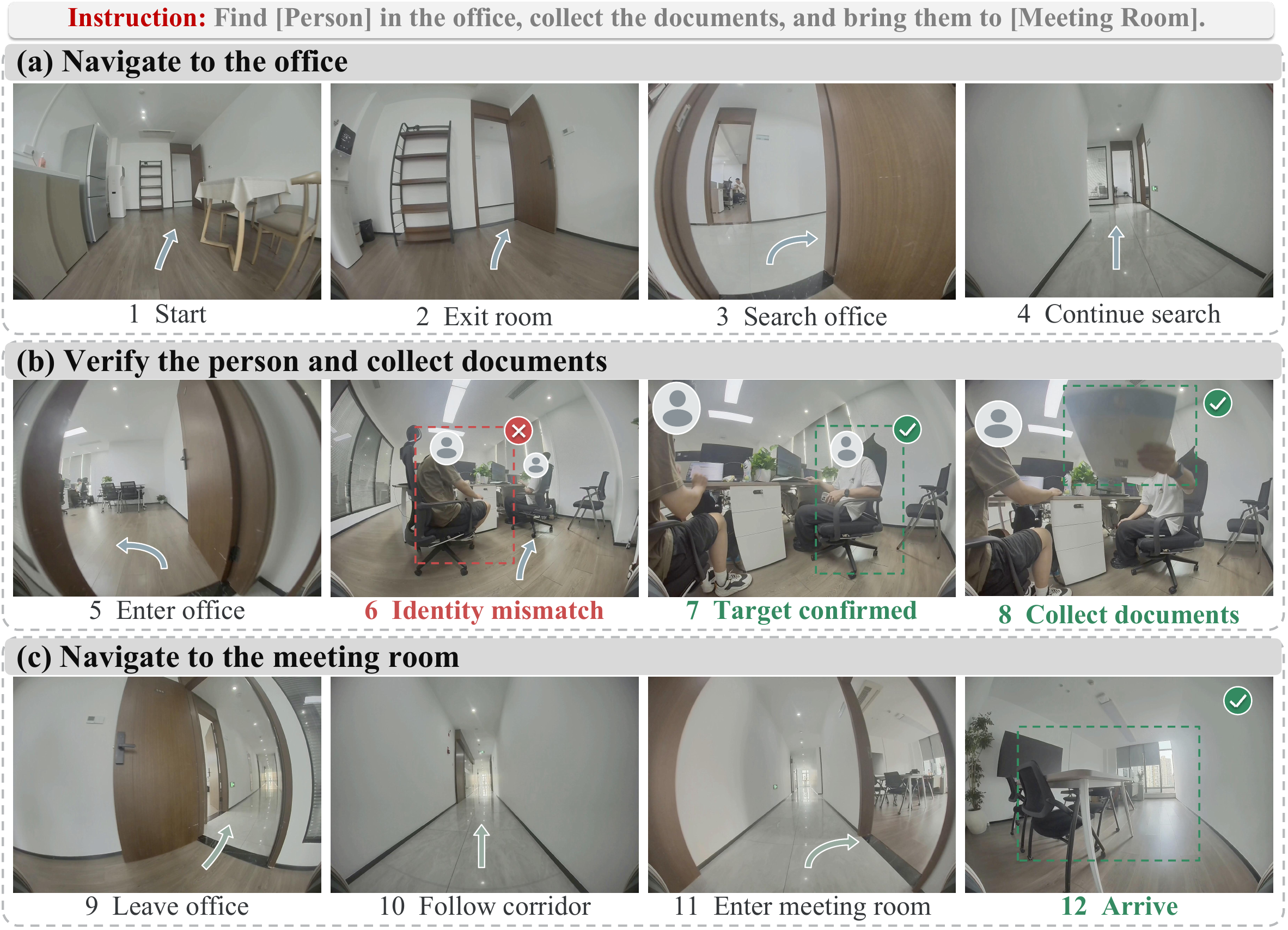}
    \caption{Long-horizon indoor document-delivery execution.}
    \label{fig:real-world-indoor-long-horizon}
\end{figure}

\begin{figure}[H]
    \centering
    \includegraphics[width=0.90\textwidth]{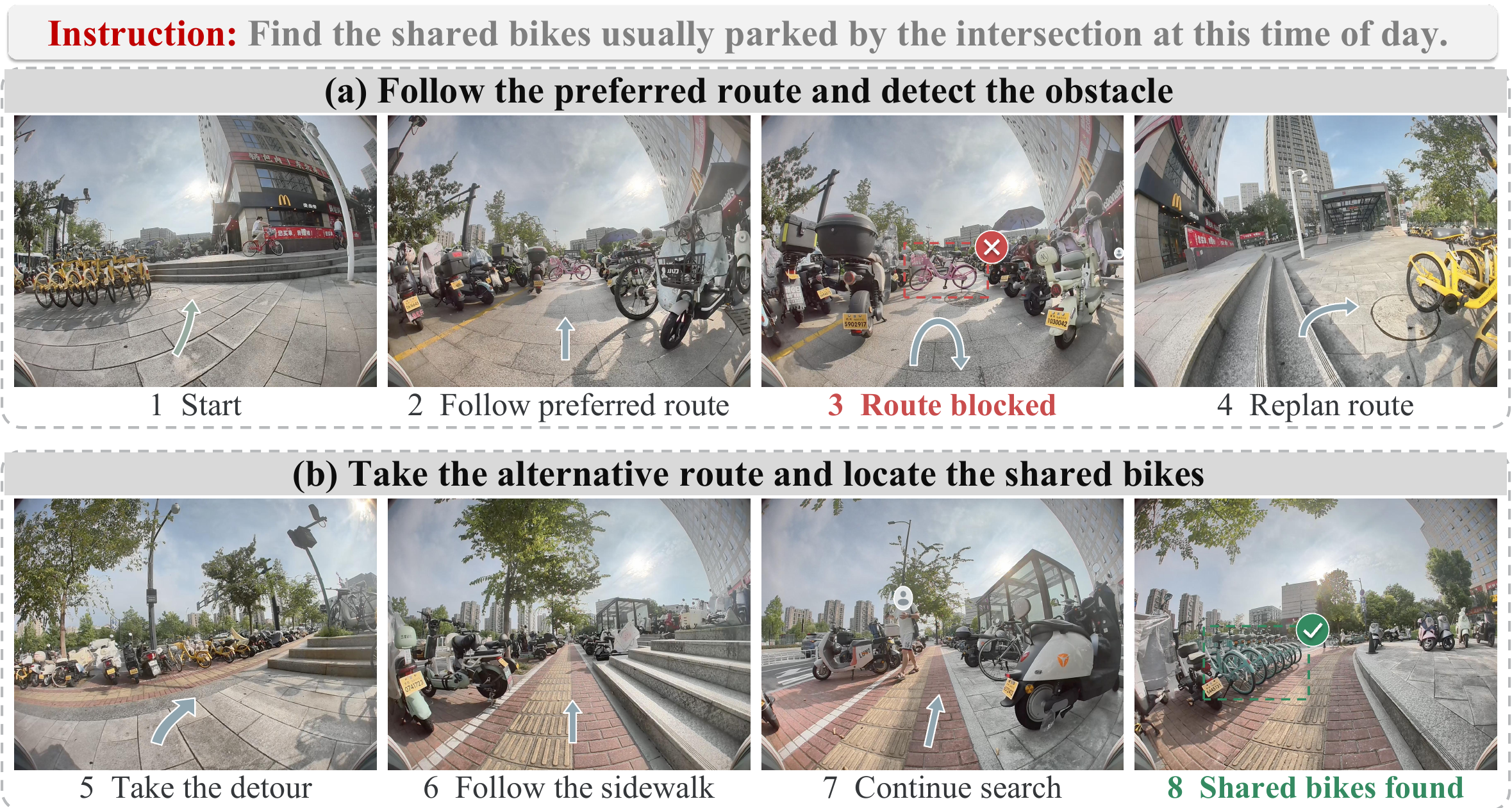}
    \caption{Long-horizon outdoor shared-bike search.}
    \label{fig:real-world-outdoor-long-horizon}
\end{figure}

\paragraph{Additional system execution cases.}
Beyond predictive target search,
\cref{fig:real-world-indoor-long-horizon,fig:real-world-outdoor-long-horizon}
document two longer-horizon executions. The indoor case combines navigation,
person verification, document collection, and delivery. The LYNX M20
has no manipulator, so a human places the documents on the robot after person
verification; the robot then autonomously transports them to the meeting room.
The outdoor case shows obstacle detection and route replanning during
shared-bike search. These cases
demonstrate broader system orchestration but are not used as quantitative
evidence for the predictive-belief module.

\FloatBarrier

\section{Reproducibility and Responsible AI}
\label{app:reproducibility}

\paragraph{Ethics and AI use.}
Persistent visual memory can capture people or private spaces, and incorrect beliefs may induce unsafe motion. Deployments should obtain consent, control data retention and access, preserve provenance, and retain platform-specific collision avoidance and emergency stops. The benchmark compiles existing records into simulated HSSD episodes without a new human-subject study. Generative AI assisted manuscript editing, LaTeX checks, and local engineering; it did not produce benchmark observations or ground-truth labels. The authors reviewed this material and remain responsible for the paper and artifacts.

\bibliographystyle{unsrtnat}
\bibliography{main}
\end{document}